\documentclass[journal]{IEEEtran}
\usepackage{amsmath,amsfonts}
\usepackage{algorithmic}
\usepackage{algorithm}
\usepackage{array}
\usepackage[caption=false, font=footnotesize]{subfig}
\usepackage{textcomp}
\usepackage{stfloats}
\usepackage{url}
\usepackage{verbatim}
\usepackage{graphicx}
\usepackage{cite}
\usepackage{cite}
\usepackage{amsmath,amsfonts}

\graphicspath{{figures/}}

\usepackage{pifont}
\usepackage{booktabs}
\usepackage{multirow}
\usepackage{rotating}
\usepackage{algorithm}
\usepackage{algorithmic}
\usepackage{capt-of}
\usepackage{url}

\newcommand{\Tref}[1]{Table~\ref{#1}}
\newcommand{\Eref}[1]{Eq.~(\ref{#1})}
\newcommand{\Fref}[1]{Fig.~\ref{#1}}

\newcommand{\Sref}[1]{Section~\ref{#1}}
\newcommand{\mymin}{\mathop{\rm {min}}\limits}

\newcommand{\cmark}{\ding{51}}%
\newcommand{\xmark}{\ding{55}}%

\begin{document}

\title{Noisy Universal Domain Adaptation \\ via Divergence Optimization for Visual Recognition}

\author{Qing~Yu,~\IEEEmembership{Student Member,~IEEE},
        Atsushi~Hashimoto,
        and~Yoshitaka~Ushiku,~\IEEEmembership{Member,~IEEE}
        % <-this % stops a space
\thanks{This study was supported by JST ACT-I Grant Number JPMJPR17U5 and partially supported by JST-Mirai Program Grant Number JPMJMI21G2 and JSPS KAKENHI Grant Number JP17H06100, Japan.}% <-this % stops a space
\thanks{Q. Yu was with the Department
of Information and Communication Engineering, The University of Tokyo, and OMRON SINIC X Corporation. E-mail: yu@hal.t.u-tokyo.ac.jp.}
\thanks{A. Hashimoto and Y. Ushiku are with OMRON SINIC X Corporation. E-mail: atsushi.hashimoto@sinicx.com, yoshitaka.ushiku@sinicx.com.}}
% \thanks{Manuscript received October 18, 2022.}}

\markboth{Journal of \LaTeX\ Class Files,~Vol.~14, No.~8, August~2021}%
{Yu \MakeLowercase{\textit{et al.}}: Noisy universal domain adaptation via divergence optimization for visual recognition}


\maketitle

\begin{abstract}
To transfer the knowledge learned from a labeled source domain to an unlabeled target domain, many studies have worked on universal domain adaptation (UniDA), where there is no constraint on the label sets of the source domain and target domain. However, the existing UniDA methods rely on source samples with correct annotations. Due to the limited resources in the real world, it is difficult to obtain a large amount of perfectly clean labeled data in a source domain in some applications. As a result, we propose a novel realistic scenario named Noisy UniDA, in which classifiers are trained using noisy labeled data from the source domain as well as unlabeled domain data from the target domain that has an uncertain class distribution. A multi-head convolutional neural network framework is proposed in this paper to address all of the challenges faced in the Noisy UniDA at once. Our network comprises a single common feature generator and multiple classifiers with various decision bounds. We can detect noisy samples in the source domain, identify unknown classes in the target domain, and align the distribution of the source and target domains by optimizing the divergence between the outputs of the various classifiers. The proposed method outperformed the existing methods in most of the settings after a thorough analysis of the various domain adaption scenarios. The source code is available at \url{https://github.com/YU1ut/Divergence-Optimization}.
\end{abstract}


% \fi
\IEEEpeerreviewmaketitle

\section{Introduction}
\label{sec:introduction}

\IEEEPARstart{W}{ith} large-scale annotated training samples, deep neural networks (DNNs) have shown outstanding results; nevertheless, when the domain of the test data is different from the domain of the training data, their performance suffers. Unsupervised domain adaptation (UDA) has been proposed to learn a discriminative classifier when there is a domain shift between the training data in the source domain and the test data in the target domain to address such distribution shifts between domains without extra annotations~\cite{ben2010theory,  french2017self, ganin2015unsupervised, ghifary2016deep, saito2017asymmetric, saito2018maximum,  saito2018maximum, sener2016learning, taigman2016unsupervised, tzeng2017adversarial}.

Although we actually do not know the class distribution of the samples in the target domain in the real world, the majority of existing domain adaptation (DA) methods assume that the source and target domains totally share the same classes. To remove the restrictions on label sets, where the target samples may contain unknown samples belonging to classes that do not exist in the source domain and some source classes may not appear in the target samples, universal DA (UniDA)~\cite{you2019universal} has been proposed.

However, current UniDA approaches require source samples with accurate annotations for model training, which means that UniDA is still an ideal situation. Because gathering clean, high-quality information is time- and money-consuming in real-world DA settings, the requirement of clean source samples restricts the applicability of existing DA methods. A crowdsourcing platform, the Internet, or social media can be used to collect data more easily; however, such data are inevitably corrupted by noise (\textit{e}.\textit{g}. YFCC100M~\cite{thomee2016yfcc100m}, Clothing1M~\cite{xiao2015learning}, and ImageNet~\cite{beyer2020we}). 

\begin{figure}[t]
    \centering
    \includegraphics[width=\linewidth]{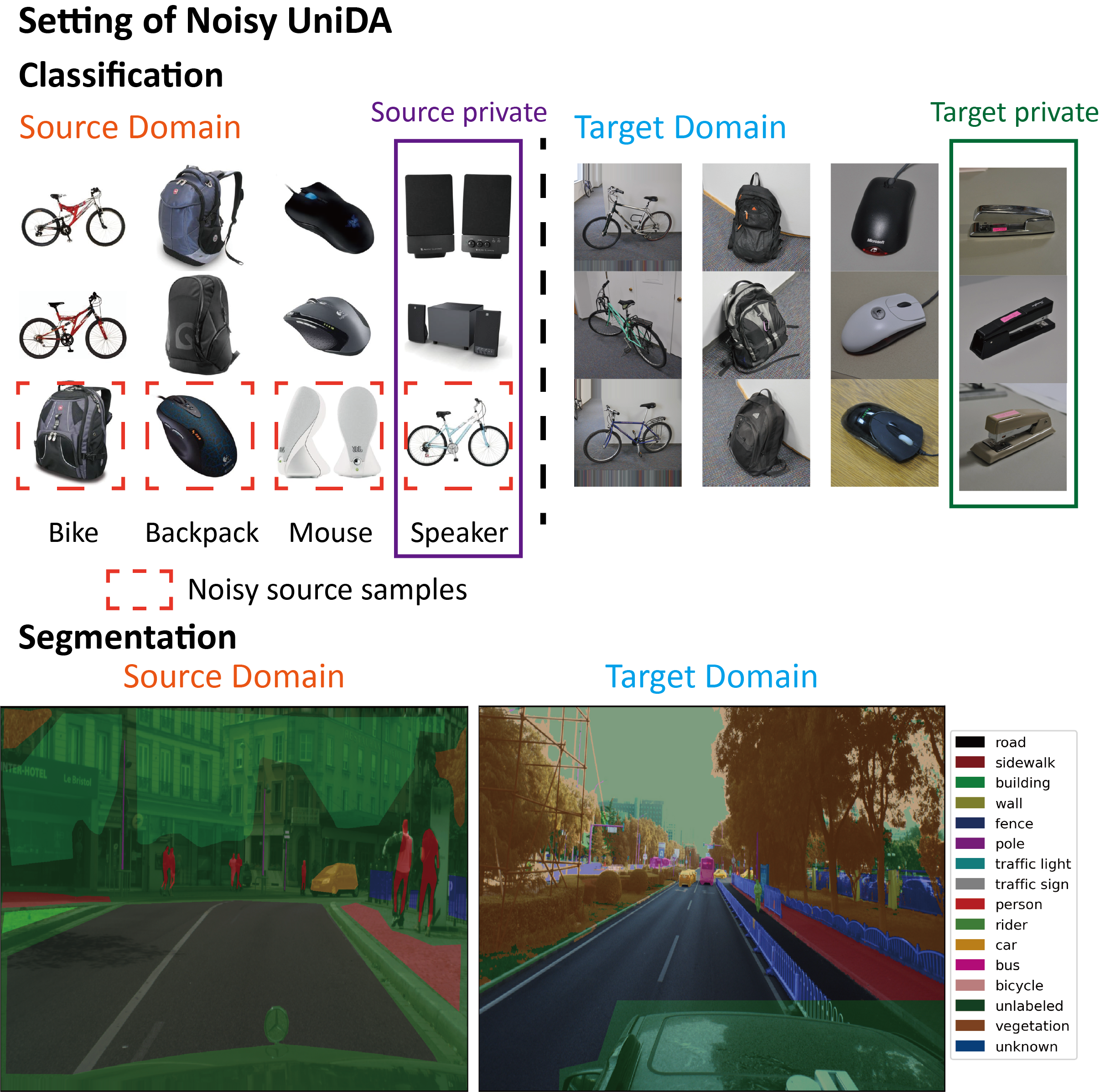}
    \caption{The proposed problem setting of Noisy UniDA, where some source samples have corrupted labels, some classes of the source domain do not appear in the target domain, and the classes of some target samples are not shared by the source domain. A novel UniDA setting with real-world label noise is also proposed for semantic segmentation.}
    \label{fig:setting}
\end{figure}

Hence, we consider a novel DA setting called ``noisy universal domain adaptation'' (Noisy UniDA), as shown in \Fref{fig:setting}. It has the following properties:
\begin{itemize}
    \setlength{\parskip}{0mm}
	\setlength{\itemsep}{0mm}
    \item Some noisy labels are contained in the labeled data in the source domain. \footnote{Because the labels of the target samples are unavailable in the UDA setting, they are not taken into account.}
    \item Some source domain classes are referred to as ``source private classes'', and they do not appear in the target domain.
    \item There are some ``target private classes'' in the target domain, which refers to some target domain classes that are not shared by the source domain.
\end{itemize}

Certain components of Noisy UniDA were solved using some of the existing methods~\cite{saito2018open,liu2019separate,shu2019transferable,cao2019learning, you2019universal}. For example, domain-adaptive models were attempted to be trained on noisy source data~\cite{shu2019transferable}, and the partial problem of source classes, where source private classes were not present in the target domain, was solved by~\cite{cao2019learning}. Additionally, the open-set problem of target private classes were addressed by~\cite{saito2018open, liu2019separate}, and the settings with the partial and open-set problems together were investigated by~\cite{you2019universal}. However, There is no solution that can address all these problems simultaneously.

To address all of the issues with Noisy UniDA at the same time, we concentrated on a unified framework of multiple classifiers with a novel proposed divergence between the classifiers. When different models with different parameters are trained on the same data, they learn distinct views of each sample because they have varied learning abilities. This technique is often used as co-training for multi-view learning and semi-supervised learning~\cite{blum1998combining, sindhwani2005co}. As a result, different models in each view will concur on the labels of the majority of samples, and compatible classifiers trained on distinct views are unlikely to concur on an incorrect label. This property can be used effectively in Noisy UniDA, where noisy source samples have inaccurate labels. Moreover, because the true labels target private samples are not included in the label set, they can be regarded as having wrong labels as well. The networks are more likely to output different predictions on these samples because they have distinct parameters. Therefore, to concurrently detect all these undesirable samples, we used the multi-head network architecture with multiple separate classifiers.

The proposed multi-head network is built with one common feature generator and multiple independent label classifiers. The separate classifiers are updated using the same data at the mini-batch level, but they are initialized differently to produce different classifiers. The divergence between the outputs of multiple classifiers on the source data was evaluated to identify noisy source samples in each mini-batch, and only the source samples with small divergences were selected to update the network using the supervised loss. At the same time, to detect target private samples and separate them from common samples, the target samples with larger divergences were selected as target private samples, and then we separated the divergence of the classifiers on the common and target private samples.  

Consequently, compared to the existing methods aligning the complete distribution, the proposed method only aligned the distributions of the clean samples from the common classes shared by both domains, which makes our method robust to inaccurate source labels, source private classes, and target private classes. We evaluated our approach with a diverse set of DA settings and our method performed significantly better than previous methods by a large margin. 

The fundamental algorithm of the proposed method and the experimental results of image classification have already been presented in our preliminary study \cite{yu2021divergence}. 
This paper extends \cite{yu2021divergence} with the following aspects:
\begin{itemize}
    \setlength\itemsep{0pt}
    \item We propose an alternative approach for multi-view learning via dropout regularization.  

    \item We extensively evaluated the classification performance in the source domain to show the effectiveness of our method in handling the noisy label problem.

    \item We are the first to tackle the problem of applying the UniDA setting to semantic segmentation including real-world label noise.
\end{itemize}

The contributions of this study are summarized as follows:
\begin{itemize}
    \setlength\itemsep{0pt}
    \item A novel experimental setting and training methodology for Noisy UniDA is proposed.  

    \item A divergence optimization framework is proposed to identify noisy source samples, detect target private samples, and achieve domain adaptation according to the divergence of multiple label classifiers.

    \item The proposed method is evaluated by the classification performance in the target and source domains across several real-world DA tasks of image classification.
    
    \item We extensively built a novel UniDA setting for semantic segmentation with real-world label noise, and our method achieved high performance.
\end{itemize}

\section{Related Studies}
There are numerous methods for UDA at the moment. The main techniques are summarized in \Tref{tbl:comp}.

There are many methods have been presented that attempt to match the distributions of the middle features in a convolutional neural network (CNN)~\cite{ganin2016domain,tzeng2014deep,saito2018maximum, zhang2019unsupervised, xu2020unsupervised, tian2021vdm}. In a domain adversarial neural network (DANN)~\cite{ganin2015unsupervised,ganin2016domain} and adversarial discriminative domain adaptation (ADDA)~\cite{tzeng2017adversarial}, a domain discriminator was trained to discriminate between two domains, while a feature extractor was trained to confuse the domain discriminator. DANN~\cite{ganin2016domain} variants have also been proposed as solutions to the issues of UniDA.

Partial DA is used when the classes of the target samples are subsets of the source classes. In studies by~\cite{cao2018partial, zhang2018importance, cao2019learning}, this problem was addressed by identifying source samples that are similar to the target samples and giving these samples more weight during the training process. When there are samples in the target domain that do not belong to a class in the source domain, the classes of these samples are named target private classes, and this setting is referred to as open-set DA. 
Previous studies have attempted to estimate the probability of acquiring an unknown class of a target sample in an adversarial learning framework to handle the open-set recognition~\cite{bendale2016towards} in a DA task. UniDA~\cite{you2019universal} was proposed in \cite{you2019universal}, and it was intended to handle partial and open-set DAs simultaneously utilizing importance weighting on both the source and target samples. Domain adaptive neighborhood clustering via entropy optimization (DANCE)~\cite{saito2020dance} uses neighborhood clustering and entropy separation to achieve weak domain alignment on the UniDA.
Although cleaning the annotations is labor- and money-intensive in practical applications, all of these methods presuppose that the annotations of the source domain are clean. Especially, it is more challenging to obtain clean annotations for the semantic segmentation task~\cite{lin2021cross, tian2021partial, lv2021weakly} because of the complicated labeling process.

The use of multiple classifiers is another popular approach for achieving DA. In the maximum classifier discrepancy (MCD)~\cite{saito2018maximum}, the source and target distributions are aligned by using task-specific decision bounds of multiple classifiers. The study~\cite{lee2019sliced} enhanced MCD~\cite{saito2018maximum} by calculating the discrepancy of multiple classifiers using a sliced Wasserstein discrepancy. Saito et al.~\cite{saito2017adversarial} also applied dropout regularization~\cite{srivastava2014dropout} to a single classifier many times to replace the multiple classifiers of MCD~\cite{saito2018maximum}. In drop to adapt \cite{lee2019drop}, the authors also apply dropout regularization \cite{srivastava2014dropout} to multiple networks multiple times for learning strongly discriminative features. Despite the fact that these methods have significantly improved the performance of the standard UDA task, they are unable to handle partial DA or open-set DA. This is because they apply strong domain alignment to all the samples, which may result in negative transfer due to the class distribution of the target domain.

\begin{table}[t]
\centering
\caption{Summary of recent related methods. UniDA consists of partial DA and open-set DA. Our proposed method is the only method that covers all the settings.}
\label{tbl:comp}
\begin{tabular}{c|c|c|c}
    \toprule
     Method & Noisy labels & Partial DA & Open-set DA  \\                 \midrule
     DANN \cite{ganin2016domain} & \xmark   & \xmark & \xmark \\
     TCL \cite{shu2019transferable} & \cmark  & \xmark & \xmark  \\ 
     ETN \cite{cao2019learning} & \xmark    & \cmark & \xmark  \\ 
     STA \cite{liu2019separate} & \xmark & \xmark & \cmark  \\ 
     UAN \cite{you2019universal} & \xmark & \cmark & \cmark  \\ 
     DANCE \cite{saito2020dance} & \xmark & \cmark & \cmark  \\ 
     Proposed & \cmark   & \cmark & \cmark  \\
    \bottomrule
\end{tabular}
\end{table}

Actually, multiple classifiers also provide the benefit of handling inaccurate annotations of source data. There have been investigations~\cite{reed2014training, zhang2016understanding, tanaka2018joint} towards the learning of discriminative models using noisy labeled datasets. By the small-loss approach, which updates the network with only the samples that have a small loss~\cite{jiang2017mentornet}, it is possible to reduce the impact of noise samples in DANN~\cite{shu2019transferable} for UDA. Another strategy~\cite{han2018co, yu2019does, wei2020combating} is to detect noisy annotations by using multiple networks to create different viewpoints of each sample and these methods have produced better outcomes.

To solve all the issues raised in Noisy UniDA, we proposed a unified framework with multiple classifiers in this study. Specifically, the proposed method is robust to source sample annotation noise levels, the situations where a subset of source classes exist in the target domain, and there are private samples in the target domain.

\section{Method}
In this section, we outline our proposed method for Noisy UniDA. First, the problem statement is defined in \Sref{sec:problem}. The overall principle of the method is then illustrated in \Sref{sec:idea}. Thereafter, the loss function is explained in \Sref{sec:loss}. Finally, the training process is described in \Sref{sec:step}.

 \subsection{Problem Statement}
 \label{sec:problem}
We suppose that a source image-label pair $\{\boldsymbol{x_{s}},\boldsymbol{y_{s}}\}$ is drawn from a set of labeled source pictures, $\{X_{s}, Y_{s}\}$, whereas $\boldsymbol{x_{t}}$, an unlabeled target image, is drawn from unlabeled images $X_{t}$. The one-hot vector of the class label $y_{s}$ is represented by $\boldsymbol{y_{s}}$. The label sets for the source and target domains are denoted by $C_s$ and $C_t$, respectively. The common label set shared by both domains is denoted by $C = C_s \cap C_t$. Additionally, we assume that the true labels for the source and target samples are $Y^{GT}_{s}$ and $Y^{GT}_{t}$ (for single source and target samples, $y^{GT}_{s}$ and $y^{GT}_{t}$ are used, respectively). This indicates that $y^{GT}_{s} \in C_s $ and $y^{GT}_{t} \in C_t$.
To deal with Noisy UniDA, the following conditions must be met in order to learn transferable features and train an accurate classifier across the source and destination domains:
\begin{itemize}
    \setlength{\parskip}{0mm}
	\setlength{\itemsep}{0mm}
    \item Noise has distorted the source image labels $Y_{s}$, which means that $\exists \{\boldsymbol{x_{s}}, y_{s}\}$, $y_{s} \neq y^{GT}_{s}$.
    \item The are some classes from the source domain that do not appear in the target domain, which means $C \subset C_s$, and these classes are denoted by $\overline{C_s} = C_s \setminus C$.
    \item There are target private samples that exist in the target domain, which means $C \subset C_t$ and these classes are denoted by $\overline{C_t} = C_t \setminus C$.
\end{itemize}

Due to the fact that the training process is carried out at the mini-batch level, $D_s = \{(\boldsymbol{x_{s}^i}, \boldsymbol{y_{s}^i})\}^{N}_{i=1}$ is indicated as a mini-batch of size $N$ sampled from the source samples, and $D_t = \{(\boldsymbol{x_{t}^i})\}^{N}_{i=1}$ is indicated as a mini-batch of size $N$ sampled from the target samples.

\subsection{Overall Concept}
\label{sec:idea}
In order to tackle the issues with Noisy UniDA, we need to train the network to accurately classify source samples under the supervision of noisy labeled source samples and align the distribution of the source and target samples while simultaneously dealing with source and target private samples.

We concentrated on the divergence of DNNs, which can solve all the issues with Noisy UniDA. Different classifiers can learn different perspectives of each sample because they can create varied decision boundaries and have different learning capacities. Additionally, they ought to be affected by the noisy labels in various ways. As a result, the outputs of the networks diverge greatly because various models are likely to concur on the labels of the majority of samples while disagreeing on the wrong labels of noisy training samples. This means that is possible to find source samples having wrong annotations and to detect target private samples that can be deemed to have inaccurate annotations because their true label does not appear in the label set.

\begin{figure}[t]
    \centering
    \includegraphics[width=8cm]{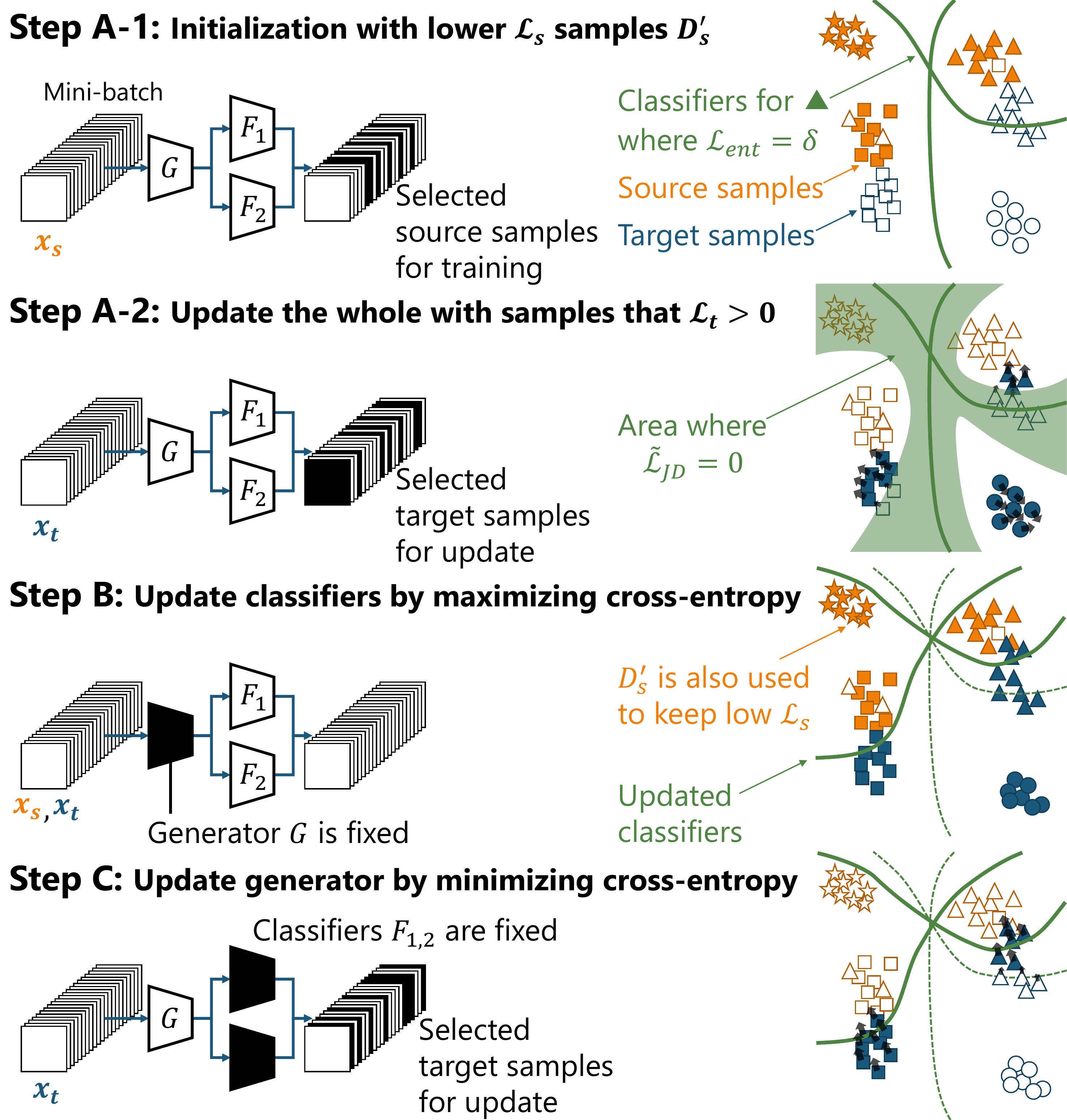}
    \caption{Training steps in the proposed method. Four classes exist in the source and target domains, and two of them are common to both domains. Unlike the filled samples, which are utilized for training, the white samples are not.}
    \label{fig:method_overview}
\end{figure}

As shown in \Fref{fig:method_overview}, we propose a divergence optimization strategy using a multi-head CNN to achieve Noisy UniDA. The multi-head CNN is made up of a feature generator network $G$ that accepts inputs of $\boldsymbol{x_{s}}$ or $\boldsymbol{x_{t}}$. The multiple classifier networks $F 1$ and $F 2$ receive features from $G$ and classify them into $|C_s|$ classes. The multiple classifiers are initialized with random initial parameters, and then they are trained using the same data at the mini-batch level. The class probabilities can be obtained by applying the softmax function to the $|C_s|$-dimensional vector of logits that is produced by the $F_1$ and $F_2$ classifier networks.

For the input $\boldsymbol{x}$, the $|C_s|$-dimensional softmax class probabilities obtained by $F_1$ and $F_2$ are denoted as $\boldsymbol{p}_1(\boldsymbol{y}|\boldsymbol{x})$ and $\boldsymbol{p}_2(\boldsymbol{y}|\boldsymbol{x})$, respectively. The probability that samples $\boldsymbol{x^i}$ belong to class $k$ predicted by each classifier  are denoted as $p_1^k(\boldsymbol{y}|\boldsymbol{x^i})$ and $p_2^k(\boldsymbol{y}|\boldsymbol{x^i})$, respectively. It is also possible to build these multiple independent classifiers by using dropout regularization~\cite{srivastava2014dropout, saito2017adversarial} repeatedly to a single classifier. \Sref{sec:adr} goes into more detail about this.

To solve the noisy label problem of source samples, we calculated the divergence between the multiple classifier outputs for each source sample at the mini-batch level. The multiple classifiers tend to output similar predictions on clean samples and different predictions on noisy data because the multiple classifiers that are trained individually have varying capacities to learn the noisy label. In each mini-batch, we chose samples with small divergences in addition to the well-known small-loss strategy to filter out the noisy samples. We further decreased the divergence of the selected clean source samples, which maximized the agreement of the multiple classifiers to achieve better results, similar to~\cite{wei2020combating}.

We propose a divergence separation loss for the target samples to address the issues with the source and target private samples. The target private samples will have larger divergences than the target common samples because they can also be thought of as noisy samples with wrong labels. Therefore, we can filter out some target private samples to achieve stable performance by separating the divergences of the target samples. Derived from the existing methods~\cite{lee2019sliced, lee2019drop, saito2017adversarial, saito2018maximum} that deploy multiple classifiers with different parameters to achieve domain adaptation, we further used the multiple classifiers as a discriminator to detect the target samples that are far from the cluster of the source domain. After that, we trained the generator to minimize the divergence to prevent it from generating target features that are not supported by the source samples.

In contrast to existing methods~\cite{lee2019sliced, lee2019drop, saito2017adversarial, saito2018maximum} that align the entire distribution of the target domain with the distribution of the source domain, \textit{we chose target samples with small divergences to update the feature generator to align the distribution partially}. Due to the partial alignment, our method can filter out the target private classes to address the issue of target private classes and focus on the samples exposed to the category boundaries to address the issue of source private classes.

To illustrate the behavior of our method, we also designed a toy problem and visualized it in \Sref{sec:toy}.

\subsection{Symmetric Kullback-Leibler and Joint Divergences}
\label{sec:loss}
The divergence between the two classifiers can be used to identify source samples having correct annotations and target private samples in the target domain, as was indicated in \Sref{sec:idea}. The symmetric Kullback-Leibler (KL) divergence, which is defined by the following equation, served as the foundation for the divergence that we used:
\begin{equation}
\mathcal{L}_{SKLD}(D_s) = {\frac{1}{N}}\sum_{i=1}^{N}D_{\mathrm{KL}}(\boldsymbol{p}_1||\boldsymbol{p}_2)+{\frac{1}{N}}\sum_{i=1}^{N}D_{\mathrm{KL}}(\boldsymbol{p}_2||\boldsymbol{p}_1),
\end{equation}
where
\begin{equation}
    D_{\mathrm{KL}}(\boldsymbol{p}_1||\boldsymbol{p}_2)=\sum_{k=1}^{|C_s|}{p_1^k(\boldsymbol{y}|\boldsymbol{x^i_s}) \log {\frac{p_1^k(\boldsymbol{y}|\boldsymbol{x^i_s})}{p_2^k(\boldsymbol{y}|\boldsymbol{x^i_s})}}},
\end{equation}
\begin{equation}
    D_{\mathrm{KL}}(\boldsymbol{p}_2||\boldsymbol{p}_1)=\sum_{k=1}^{|C_s|}{p_2^k(\boldsymbol{y}|\boldsymbol{x^i_s}) \log {\frac{p_2^k(\boldsymbol{y}|\boldsymbol{x^i_s})}{p_1^k(\boldsymbol{y}|\boldsymbol{x^i_s})}}}.
\end{equation}

When it comes to the source classes, we directly used $\mathcal{L}_{SKLD}$ to measure the agreement of the classifiers to identify the clean source samples. We then minimized $\mathcal{L}_{SKLD}$ \textit{on these clean source samples}.

While taking into account the target classes, we try to use the divergence to find samples belonging to the target private classes. According to our analysis, the expression of $D_{\mathrm{KL}}(\boldsymbol{p}_1||\boldsymbol{p}_2)$ can be rewritten as follows:

\begin{eqnarray}
\begin{aligned}
    D_{\mathrm{KL}}(\boldsymbol{p}_1||\boldsymbol{p}_2)=&\sum_{k=1}^{|C_s|}{p_1^k(\boldsymbol{y}|\boldsymbol{x^i_t}) \log {p_1^k(\boldsymbol{y}|\boldsymbol{x^i_t})}}
\\ &-
\sum_{k=1}^{|C_s|}{p_1^k(\boldsymbol{y}|\boldsymbol{x^i_t}) \log {p_2^k(\boldsymbol{y}|\boldsymbol{x^i_t})}}
\\ = -H(\boldsymbol{p}_1(\boldsymbol{y}|\boldsymbol{x_t}))&+H(\boldsymbol{p}_1(\boldsymbol{y}|\boldsymbol{x_t}),\boldsymbol{p}_2(\boldsymbol{y}|\boldsymbol{x_t})),
\end{aligned}
\end{eqnarray}
where $H(\boldsymbol{p}_1(\boldsymbol{y}|\boldsymbol{x_t}))$ is the entropy of $\boldsymbol{p}_1(\boldsymbol{y}|\boldsymbol{x_t})$, and $H(\boldsymbol{p}_1(\boldsymbol{y}|\boldsymbol{x_t}),\boldsymbol{p}_2(\boldsymbol{y}|\boldsymbol{x_t}))$ is the cross-entropy for $\boldsymbol{p}_1(\boldsymbol{y}|\boldsymbol{x_t})$ and $\boldsymbol{p}_2(\boldsymbol{y}|\boldsymbol{x_t})$.

Therefore, $\mathcal{L}_{SKLD}$ can be rewritten as:
\begin{eqnarray}
\begin{aligned}
\label{eq:kl}
\mathcal{L}_{SKLD}(D_t) &= {\frac{1}{N}}\sum_{i=1}^{N}\mathcal{L}_{crs}(D_t)-{\frac{1}{N}}\sum_{i=1}^{N}\mathcal{L}_{ent}(D_t)
\\ \mathcal{L}_{crs}(D_t) &= H(\boldsymbol{p}_1(\boldsymbol{y}|\boldsymbol{x_t}),\boldsymbol{p}_2(\boldsymbol{y}|\boldsymbol{x_t}))\\&+H(\boldsymbol{p}_2(\boldsymbol{y}|\boldsymbol{x_t}),\boldsymbol{p}_1(\boldsymbol{y}|\boldsymbol{x_t}))
\\\mathcal{L}_{ent}(D_t) &=H(\boldsymbol{p}_1(\boldsymbol{y}|\boldsymbol{x_t}))+H(\boldsymbol{p}_2(\boldsymbol{y}|\boldsymbol{x_t})),
\end{aligned}
\end{eqnarray}
where the first and second terms show the divergence of the outputs of the two classifiers and the entropy of each classifier output, respectively.

We noted in \Sref{sec:idea} that target private samples would probably have larger divergences than target common samples. The class probabilities of the target private samples will have a ``small'' entropy if we directly apply the symmetric KL divergence to assess the divergence because the second term in \Eref{eq:kl} has a minus symbol. But in practice, the prediction confidence of these samples should be low, as the target private samples do not belong to any source classes, which suggests that the entropy of their class probabilities should be ``high''.

As a result, to find the target private samples, we changed the symmetric KL divergence to ``Joint Divergence'' as follows:
\begin{equation}
\label{eq:kl_mod}
    \mathcal{L}_{JD}(D_t) = {\frac{1}{N}}\sum_{i=1}^{N}\mathcal{L}_{crs}(D_t) + {\frac{1}{N}}\sum_{i=1}^{N} \mathcal{L}_{ent}(D_t),
\end{equation}
where a larger divergence denotes lower confidence in each prediction and a larger degree of disagreement between the two classifiers. Additionally, for the detected target common samples and the detected target private samples, we minimized and maximized \Eref{eq:kl_mod}, respectively. The following section goes into more details on the training process.

\subsection{Training Procedure}
\label{sec:step}
Taking into account the previous discussions in \Sref{sec:idea} and \Sref{sec:loss}, we propose a novel training procedure with three steps, as illustrated in \Fref{fig:method_overview}. \textit{In our method, the three steps are repeated at the mini-batch level.}

\textbf{Step A-1}
In order to categorize the source samples correctly, we first trained the complete network, which includes both classifiers and the generator, to learn discriminative features under the supervision of the labeled source samples. We trained our classifier using only small-loss instances in each mini-batch of data in order to make the network robust to noisy labels because small-loss samples were likely to have correct labels~\cite{han2018co, wei2020combating}. Commonly used as the loss function, cross-entropy loss is represented as follows:
\begin{equation}
\begin{aligned}
   \mathcal{L}_{sup}(D_s) = &-{\frac{1}{N}}\sum_{i=1}^{N}\sum_{k=1}^{|C_s|}y_s^i\log p_1^k({\boldsymbol y}|{\boldsymbol {x^i_s}})
   \\&-{\frac{1}{N}}\sum_{i=1}^{N}\sum_{k=1}^{|C_s|}y_s^i\log p_2^k({\boldsymbol y}|{\boldsymbol {x^i_s}}).
\end{aligned}
\label{eq:l_sup}
\end{equation}

We also incorporated the agreement between the two classifiers to the loss to choose clean samples, as was indicated in \Sref{sec:loss}. Consequently, the following is how the loss of the source samples is expressed:
\begin{equation}
\label{eq:Ls}
\begin{aligned}
   \mathcal{L}_{s}(D_s) = \mathcal{L}_{sup}(D_s) +\lambda \mathcal{L}_{SKLD}(D_s),
\end{aligned}
\end{equation}
where the hyperparameter $\lambda$ is always set to 0.1 in the experiments.
Because a noisy sample was more likely to have a larger cross-entropy loss and larger divergence, we deployed the joint loss \Eref{eq:Ls} to filter out the noisy samples. To be more precise, we performed a small-loss selection as follows:
\begin{equation}
\label{eq:select}
D_s'={\arg\min}_{D':|D'|\ge \alpha|D_s|}\mathcal{L}_{s}(D_s).
\end{equation}

This equation shows that we only used $\alpha\%$ samples in the mini-batch. The following are the objectives:
\begin{equation}
  \mymin_{G,F_1,F_2} \mathcal{L}_s(D_s').
\end{equation}

\textbf{Step A-2}
Besides the training with the source samples, we attempted to use $\mathcal{L}_{JD}$ from \Eref{eq:kl_mod} to identify the target private samples. We added a threshold $\delta$ and a margin $m$ to separate the $\mathcal{L}_{JD}$ on target samples in order to increase $\mathcal{L}_{JD}$ for the target private samples and decrease it for the target common samples. This separation operation is denoted as follows:
\begin{equation}
\label{eq:Lt}
\begin{aligned}
   &\mathcal{L}_{t}(D_t) = \mathcal{\tilde{L}}_{JD}(D_t)={\frac{1}{N}}\sum_{i=1}^{N}\mathcal{\tilde{L}}_{crs}(D_t)+{\frac{1}{N}}\sum_{i=1}^{N}\mathcal{\tilde{L}}_{ent}(D_t)
   \\&\mathcal{\tilde{L}}_{crs}(D_t)=
   \begin{cases}
   -|\mathcal{L}_{crs}(\boldsymbol {x_t})-\delta| & \text{if } |\mathcal{L}_{crs}(\boldsymbol {x_t})-\delta| > m\\
   0 & \text{otherwise}
   \end{cases}
   \\&\mathcal{\tilde{L}}_{ent}(D_t)=
   \begin{cases}
   -|\mathcal{L}_{ent}(\boldsymbol {x_t})-\delta| & \text{if } |\mathcal{L}_{ent}(\boldsymbol {x_t})-\delta| > m\\
   0 & \text{otherwise}
   \end{cases}.
\end{aligned}
\end{equation}

Only when the divergence or entropy of the two classifiers was sufficiently large or small, did we further increase or decrease them to separate private or common target samples. The following are the objectives:
\begin{equation}
  \mymin_{G,F_1,F_2} \mathcal{L}_{t}(D_t).
\end{equation}

In addition, $D_t'$ represents the common target sample with small divergence, which is $\{\boldsymbol{x_{t}^i}: \boldsymbol{x_{t}^i} \in D_t, \mathcal{L}_{crs}(\boldsymbol{x_{t}^i}) < \delta - m\} $.

\textbf{Step B}
The classifiers were then trained as a discriminator for a fixed generator to increase the divergence of the outputs and detect target samples that lack the support of the source samples in order to align the distribution of the source and target domains (\textbf{Step B} in \Fref{fig:method_overview}). In this step, we additionally reshaped the support using source samples. The following is the objective:
\begin{equation}
\label{eq:step_b}
  \mymin_{F_1,F_2} \mathcal{L}_s(D_s') - {\frac{1}{N}}\sum_{i=1}^{N}\mathcal{L}_{crs}(D_t). \\
\end{equation}
  
\textbf{Step C}
Finally, using the identified target common samples $D_t'$, we trained the generator to minimize the divergence of fixed classifiers (\textbf{Step C} in \Fref{fig:method_overview}) to partially align the distributions of the target and source domains. The final objective is as follows:
\begin{equation}
\label{eq:step_c}
 \mymin_{G} {\frac{1}{N}}\sum_{i=1}^{N}\mathcal{L}_{crs}(D_t'). \\
 \end{equation}
 
We repeat this step $n$ times to improve the domain alignment and $n=4$ is used in all the experiments.

\subsection{Inference}
\label{sec:test}
At the inference time, we took into account the cross entropy $\mathcal{L}_{crs}$ between the outputs of the two classifiers to discriminate between the common and target private samples. The sample is designated as a target private sample when the divergence is larger than the detection threshold $\delta$ and is indicated by:
\begin{equation}
\label{eq:inf}
    \mathcal{L}_{crs}(\boldsymbol{x}) > \delta.
\end{equation}

\subsection{Alternative Approach via Dropout Regularization }
\label{sec:adr}
As mentioned in \Sref{sec:idea}, we used two different classifiers to achieve multi-view learning for domain adaptation. However, using two classifiers introduces additional parameters to the network. This may limit the application of this method in some cases. To solve this problem, we applied dropout regularization \cite{srivastava2014dropout, saito2017adversarial} to a single network that can also achieve a comparative performance. 

When a neural network is trained with dropout, each node of the network is removed with some probability for each sample within a mini-batch. This operation is equal to selecting a different classifier for each sample during the training process. Consequently, if we forward the features generated by $G$ to the classifier $F$ twice by dropping different nodes each time, we can obtain two different output vectors, $\boldsymbol{p}_1(\boldsymbol{y}|\boldsymbol{x})$ and $\boldsymbol{p}_2(\boldsymbol{y}|\boldsymbol{x})$. Alternatively, we used dropout to select two different classifiers, $F_1$ and $F_2$, from $F$.

\section{Experiments on Toy Datasets}
\label{sec:toy}
To build a toy dataset, we used scikit-learn to create isotropic Gaussian blobs as the source and target samples, and we observed the behavior of the proposed method. The goal of the experiment was to observe the boundaries of classifiers after the training. We generated three clusters of blobs and assigned them the colors red, blue, and orange as the source samples. The source samples also contained label noise. Target private samples were generated far from the source samples, whereas target common samples were generated in the area where the red and blue source samples were distributed. As the training samples, we created 300 source and target samples for each class.

A feature generator network with three fully-connected layers and the classifiers with fully-connected layers make up the model. We illustrate the learned decision boundary of the proposed method and its variations as follows.

\textbf{Source Only} 
The variation that trained with source samples only with \Eref{eq:l_sup}, whose final objective is denoted as:
\setcounter{equation}{15}
\begin{equation}
  \mymin_{G,F_1,F_2} \mathcal{L}_{sup}(D_s).
\end{equation}

\textbf{Ours w/o select} 
The variation that trained without using the small-loss selection by \Eref{eq:select}, which means that the objective of Step A-1 is as follows:
\begin{equation}
  \mymin_{G,F_1,F_2} \mathcal{L}_s(D_s).
\end{equation}

\textbf{Ours w/o sep} 
The variation that trained without separating the divergence between the classifiers by \Eref{eq:Lt}, which means that the following equation is used instead of \Eref{eq:Lt}:
\begin{equation}
  \mathcal{L}_{t}(D_t) = 0.
\end{equation}

\textbf{Ours w/ KL} 
The variation that trained by using the general symmetric KL-divergence in \Eref{eq:Lt}, which means that the following equation is used instead of \Eref{eq:Lt}:
\begin{equation}
   \mathcal{L}_{t}(D_t) = \mathcal{\tilde{L}}_{SKLD}(D_t)={\frac{1}{N}}\sum_{i=1}^{N}\mathcal{\tilde{L}}_{crs}(D_t)-{\frac{1}{N}}\sum_{i=1}^{N}\mathcal{\tilde{L}}_{ent}(D_t)
\end{equation}

\begin{figure*} 
    \centering
  \subfloat[Source only\label{fig:toy_so}]{%
       \includegraphics[width=0.33\linewidth]{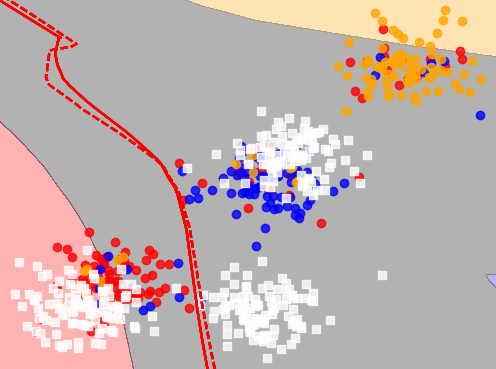}}
    \hfill
  \subfloat[Ours w/o select\label{fig:toy_sel}]{%
        \includegraphics[width=0.33\linewidth]{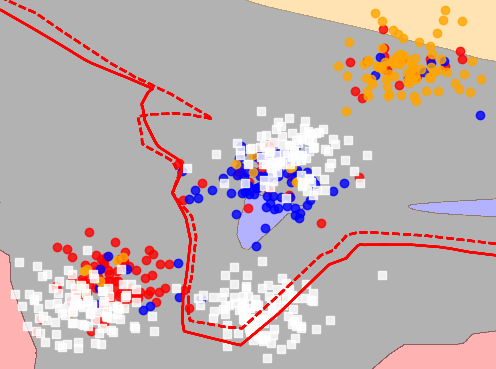}}
    \hfill
  \subfloat[Ours w/o sep\label{fig:toy_sep}]{%
        \includegraphics[width=0.33\linewidth]{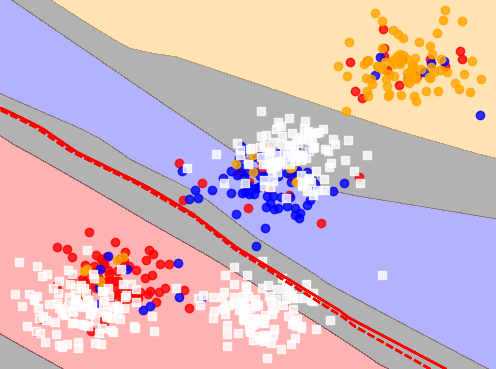}}
    \hfill
  \subfloat[Ours w/ KL\label{fig:toy_kl}]{%
        \includegraphics[width=0.33\linewidth]{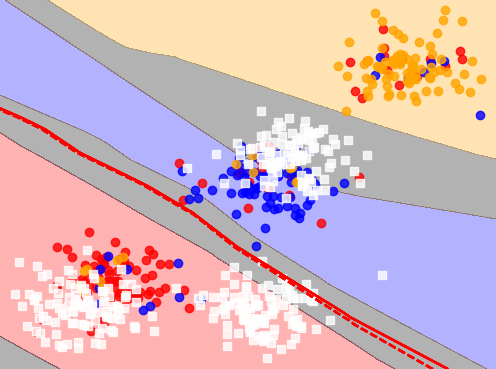}}
  \hspace*{1pt}
  \subfloat[Ours\label{fig:toy_ours}]{%
        \includegraphics[width=0.33\linewidth]{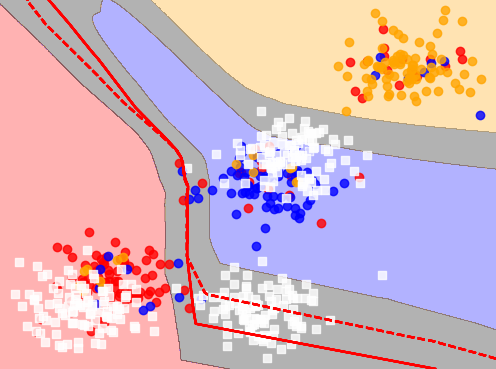}}
  \caption {(Best viewed in color.) The three classes of the source samples are represented by red, blue, and orange points, with the orange class serving as the source private class. 20\% symmetric noise has contaminated the labels of the source samples. The target samples are shown as white points, and the target samples in the bottom right corner of each figure are the target private samples. The two decision boundaries in our method for determining the red class are represented by the dashed and normal lines. The output of both classifiers falls into the categories of red, blue, and orange, respectively, in the pink, light blue, and light yellow regions. By using \Eref{eq:inf}, the target private classes with a large divergence are identified in the gray regions.}
\label{fig:toy}
\end{figure*}

Due to the label noise of the source samples, a significant region is identified as the target private class with a large $\mathcal{L}_{crs}$ in the source only and Ours w/o select as shown in \Fref{fig:toy_so} and \Fref{fig:toy_sel} respectively. Even though Ours w/o sep and Ours w/ KL shown in \Fref{fig:toy_sep} and \Fref{fig:toy_kl} are unaffected by label noise, they struggle to pick up on target private samples. This is due to the fact that their loss function is inadequate to distinguish between the target common and private classes. The proposed method demonstrated in \Fref{fig:toy_ours} achieved the best performance when compared to other variations. The classifiers are unaffected by the noisy source samples when using our proposed method, and target private samples exist in the region of large divergence. We aim to make these target private samples more divergent than the target common samples.
	
\section{Experiments on Real-world Datasets}
\label{sec:exp}
In this section, our experimental settings and results on real-world datasets are discussed.
\subsection{Experimental Setup}
\subsubsection{Datasets}
We used three datasets for the experiments, which are the same as the previous studies~\cite{you2019universal}. The first dataset is Office~\cite{saenko2010adapting}, which has 31 classes and three domains (Amazon, DSLR, and Webcam). The second dataset is OfficeHome~\cite{venkateswara2017deep}, which has 65 classes and four domains (Real, Product, Clipart, and Art). The final dataset is VisDA~\cite{peng2017visda}, which has 12 classes and two domains (Synthetic and Real).
The classes of each dataset were divided as in~\cite{you2019universal} to create the Noisy UniDA setting: $|C|/|\overline{C_s}|/|\overline{C_t}|=10/10/11$ for Office, $10/5/50$ for OfficeHome, and $6/3/3$ for VisDA. In order to simulate the noisy source samples, we also manually corrupted the source datasets using a noise transition matrix $Q$~\cite{han2018co,jiang2017mentornet}. We implemented two versions of $Q$: one is pair flipping and the other one is symmetry flipping \cite{han2018co}. The range of $\{0.2,0.45\}$ was used to select the noise rate, $\rho$. According to intuition, around half of the noisy source data had inaccurate labels that could not have been learned without additional presumptions when $\rho=0.45$.
When just $20\%$ of the labels are thought to be distorted by $\rho=0.2$, this represents a low-level noise problem. According to~\cite{han2018co}, pair flipping is more challenging than symmetry flipping. There were four different types of noisy source data for each adaptation task: \emph{Pair-}$20\%$ (P$20$), \emph{Pair-}$45\%$ (P$45$), \emph{Symmetry-}$20\%$ (S$20$),  and \emph{Symmetry-}$45\%$ (S$45$).

\begin{table*}
\begin{minipage}{0.6\linewidth}
\centering
\caption{Average target-domain accuracy (\%) of each dataset under different noise types. We report the average accuracy of all the tasks for each dataset. The bold values represent the highest accuracy for each row.}
\scalebox{0.9}{
\tabcolsep = 0.7mm
\begin{tabular}{c|cccc|cccc|cccc}
\toprule
\multirow{2}{*}{Method} & \multicolumn{4}{c|}{Office} & \multicolumn{4}{c|}{OfficeHome} & \multicolumn{4}{c}{VisDA} \\ \cline{2-13}
                        & P20   & P45  & S20  & S45  & P20    & P45   & S20   & S45   & P20  & P45  & S20  & S45  \\ \hline\hline
SO                      &   77.23    &  50.88    &  78.09    &  53.15    &    63.33    &   39.46    &  64.09     &   44.99    &   44.57   &   38.41   &  26.51    &   15.39   \\
TCL                     &   80.82    &  50.48    & 82.97     &  74.86    &   62.31     &  40.24     & 63.41      &   49.69    &  62.96    &   43.31   &  61.26    &   52.96   \\
ETN                     &   84.46    &  53.23    &  85.40    &  83.53    &    67.93    &  44.55     &  68.99     &   56.05    &   58.99   &   44.36   &  62.17    &   55.83   \\
STA                     &   83.12    &   54.74   &  83.19    & 68.27     &    64.31    &  44.22     &  65.53     &    49.04   &    41.62  &  41.50    &  52.17    &  42.32    \\
UAN                     &   72.39    &  45.64    &   77.59   &  64.85    &    70.90    &  41.31     &  73.79     &    59.67   &   53.93   &  42.60    &  53.25    &   47.70   \\
DANCE                   &   84.88    &  55.40     &  83.00    &  56.02    &    \textbf{77.32}    &  47.51     &   77.54    &   66.96    &   57.38   &  41.45    &  24.30    &   14.94   \\
DANCE$_{\text{select}}$   &   86.07    &  57.87     &  91.24    &  79.82  &  76.49    &  48.45     &   78.71    &  64.17 &  63.93   &  43.91    &  62.97    &   52.32   \\ \hline
Ours                    &   91.22    &  \textbf{62.49}    &  91.40    &  87.92    &   76.10     &  51.93     &   77.46    &  71.97     &   67.27   &  \textbf{48.25}    &   70.53   & 57.82  \\
Ours$_{\text{dropout}}$  &   \textbf{91.79}    &  62.32    &  \textbf{93.04}    &  \textbf{89.70}    &   76.70     &  \textbf{52.28}     &   \textbf{79.41}    &  \textbf{74.47}     &   \textbf{73.86}   &  46.35    &   \textbf{75.04}   & \textbf{71.75} \\ \bottomrule
\end{tabular}
}
\label{tbl:uni_results}
\end{minipage}
\begin{minipage}{0.4\linewidth}
		\centering
		\includegraphics[width=0.85\textwidth]{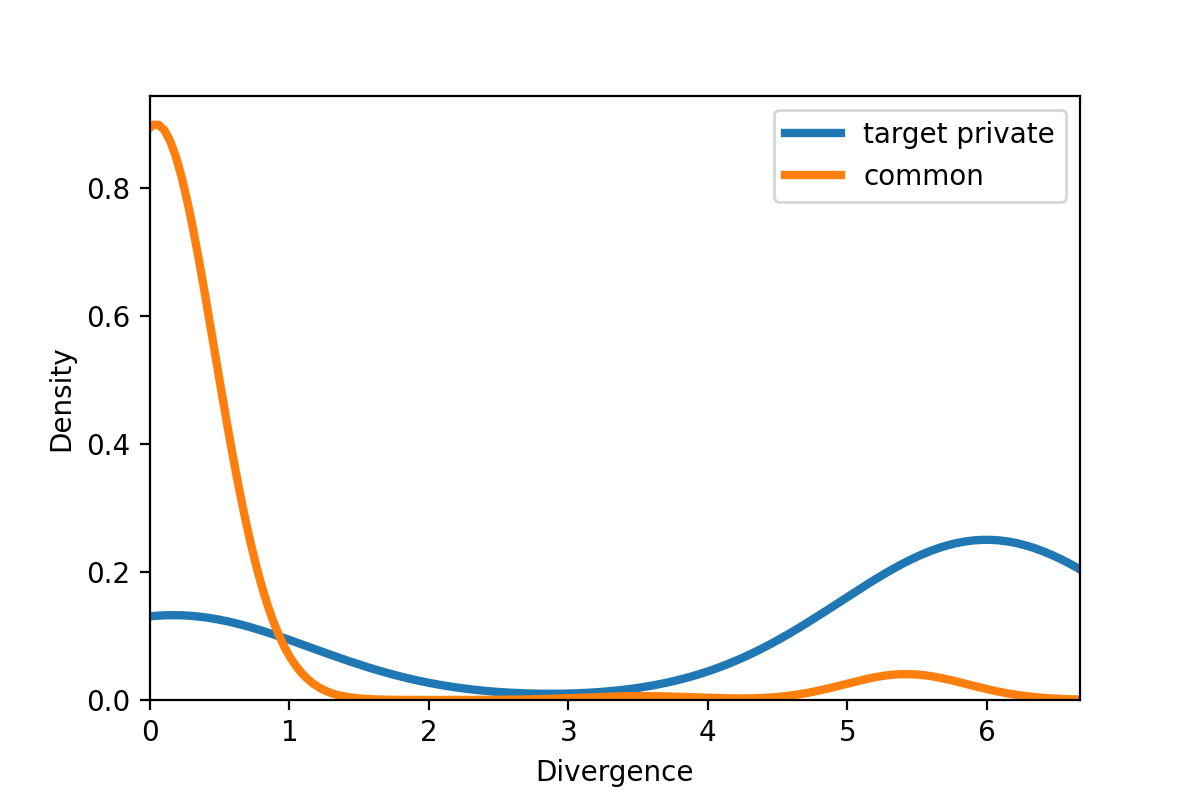}
	    \vspace{-5pt}
		\captionof{figure}{Probability density function of the divergence of common and target private samples (estimated using Gaussian kernel with Scott's rule).}
		\label{fig:dis}
	\end{minipage}
\end{table*}

\begin{table*}[t]
\centering
\caption{Results of each task on noisy universal domain adaptation for Office and OfficeHome. The bold values represent the highest accuracy for each row.}
\tabcolsep = 0.7mm
\begin{tabular}{c|cccccc|c|cccccccccccc|c}
\multicolumn{21}{c}{Noise Type: P45}  \\
\toprule
\multirow{2}{*}{Method} & \multicolumn{6}{c|}{Office}        &     & \multicolumn{12}{c|}{OfficeHome}                                       &     \\ \cline{2-21}
                        & A2W & D2W & W2D & A2D & D2A & W2A & Avg & A2C & A2P & A2R & C2A & C2P & C2R & P2A & P2C & P2R & R2A & R2C & R2P & Avg \\ \hline\hline
SO &47.12&56.50&59.13&50.67&56.00&35.89&50.88&33.11&41.16&57.84&30.01&37.50&41.86&39.30&26.03&50.57&40.81&28.51&46.80&39.46     \\
TCL  &50.29&54.97&52.85&49.58&50.77&44.41&50.48&29.40&42.62&51.52&31.73&39.60&42.79&50.39&28.20&53.34&34.65&30.27&48.34&40.24    \\
ETN    &52.01&52.94&49.47&52.79&60.33&51.82&53.23&36.36&54.08&\textbf{65.48}&35.36&38.74&49.39&47.04&27.04&57.09&41.50&33.35&49.15&44.55    \\
STA  &61.18&53.42&54.19&55.93&63.05&40.69&54.74&32.21&42.02&60.62&38.38&42.37&53.13&50.29&30.48&58.99&42.47&28.60&51.03&44.22   \\
UAN &44.93&57.68&44.00&41.09&40.50&45.64&45.64&31.53&44.44&46.64&40.16&44.04&47.48&41.42&34.38&54.81&38.61&30.06&42.15&41.31  \\
DANCE  &46.80&56.82&53.85&56.17&\textbf{69.98}&48.80&55.40&36.10&39.69&63.12&39.62&41.60&46.84&57.04&32.28&68.55&50.26&39.82&55.20&47.51    \\
DANCE$_{\text{select}}$  &65.08&60.32&\textbf{68.14}&48.79&54.42&50.48&57.87&34.32&41.95&64.19&47.25&42.24&\textbf{61.73}&46.32&40.92&60.40&53.30&33.86&54.87&48.45  \\ \hline
Ours  &58.93&\textbf{72.49}&56.06&58.71&65.86&\textbf{62.90}&\textbf{62.49}&\textbf{37.12}&57.73&54.17&\textbf{52.39}&\textbf{47.26}&55.22&\textbf{57.93}&\textbf{43.97}&64.17&\textbf{50.36}&41.29&61.61&51.93 \\   
Ours$_{\text{dropout}}$&\textbf{68.08}&66.68&57.10&\textbf{59.78}&62.28&60.02&62.32&35.35&\textbf{60.79}&59.48&43.86&46.71&55.45&47.52&40.17&\textbf{68.91}&48.21&\textbf{53.77}&\textbf{67.17}&\textbf{52.28}\\\bottomrule
\multicolumn{21}{c}{}  \\
\multicolumn{21}{c}{Noise Type: S45}  \\
\toprule
\multirow{2}{*}{Method} & \multicolumn{6}{c|}{Office}        &     & \multicolumn{12}{c|}{OfficeHome}                                       &     \\ \cline{2-21}
                        & A2W & D2W & W2D & A2D & D2A & W2A & Avg & A2C & A2P & A2R & C2A & C2P & C2R & P2A & P2C & P2R & R2A & R2C & R2P & Avg \\ \hline\hline
SO &37.16&53.17&76.00&50.49&41.97&60.14&53.15&30.36&46.40&59.20&48.84&46.50&57.94&41.20&29.09&56.49&40.56&30.39&52.90&44.99    \\
TCL  &77.62&64.50&82.96&81.86&66.35&75.84&74.86&30.08&44.22&46.66&49.31&54.73&57.03&52.26&40.72&69.14&48.61&40.41&63.16&49.69    \\
ETN   &83.11&80.64&88.64&87.35&77.18&84.27&83.53&39.96&62.37&77.07&61.08&53.51&70.17&47.70&39.42&68.93&49.77&37.32&65.33&56.05    \\
STA  &60.94&70.22&88.80&71.6&44.26&73.79&68.27&37.65&50.75&57.14&51.35&53.64&67.39&39.92&32.09&64.29&47.19&32.63&54.49&49.04  \\
UAN &62.59&72.88&75.82&56.97&61.94&58.90&64.85&42.67&59.04&57.20&57.29&64.93&70.56&64.45&49.09&72.53&55.98&47.26&75.08&59.67  \\
DANCE  &21.72&69.45&85.76&27.68&51.04&80.47&56.02&38.29&\textbf{74.72}&87.24&\textbf{75.01}&\textbf{81.27}&80.20&55.57&44.05&78.08&61.59&47.70&79.83&66.96    \\ 
DANCE$_{\text{select}}$ &82.45&69.16&92.08&86.53&75.51&73.22&79.82&40.33&58.27&73.80&61.15&67.76&78.36&70.63&54.23&78.53&63.22&48.21&75.56&64.17 \\\hline
Ours &87.63&76.87&\textbf{98.32}&\textbf{89.43}&\textbf{84.49}&90.78&87.92&46.30&69.52&\textbf{87.79}&70.34&70.55&81.77&74.72&54.15&\textbf{88.23}&78.04&61.22&80.98&71.97 \\  
Ours$_{\text{dropout}}$ &\textbf{89.48}&\textbf{87.19}&98.11&88.97&83.65&\textbf{90.81}&\textbf{89.70}&\textbf{54.72}&72.26&81.07&74.46&72.94&\textbf{86.59}&\textbf{78.05}&\textbf{55.87}&88.07&\textbf{80.87}&\textbf{64.49}&\textbf{84.28}&\textbf{74.47} \\   \bottomrule
\end{tabular}
\label{tbl:detail_results}
\end{table*}

\subsubsection{Evaluation Details}
\textbf{Compared Methods} We compared the proposed method to the following approaches: (1) CNN: source only ResNet-50 (SO)~\cite{he2016deep}; (2) label noise-tolerant domain adaptation method: TCL~\cite{shu2019transferable}; (3) partial domain adaptation method: example transfer network (ETN)~\cite{zhang2018importance}; (4) open-set domain adaptation method: separate to adapt (STA)~\cite{liu2019separate}; and (5) universal domain adaptation methods: universal adaptation network (UAN)~\cite{you2019universal} and DANCE~\cite{saito2020dance}. It is important to demonstrate how these approaches perform in the Noisy UniDA setting because they have attained state-of-the-art performance in their individual settings.

\textbf{Evaluation Protocols} The same evaluation metrics that were applied in existing methods are used here. All samples corresponding to the target private classes are treated as a single, unified unknown class, and the accuracy is averaged over $|C|+1$ classes. For example, an average of 11 classes is provided when the Office dataset is used. We deployed confidence thresholding to identify the target private samples for the methods that initially did not detect them.

\textbf{Implementation Details}
We deployed the same CNN architecture and hyperparameters for this experiment as in~\cite{saito2020dance}. Our network was based on ResNet-50~\cite{he2016deep}. The modules of ResNet until the {\it average-pooling} layer just before the last {\it full-connected} layer were used as the generator and one {\it full-connected} layer were used as the classifier. Because $2 \times \log |C_s|$ is the maximum value of $H(\boldsymbol{p_1})+H(\boldsymbol{p_2})$, we defined the threshold as $\delta = \log |C_s|$ and the margin $m=1$ in all the experiments. 5,000 iterations were used to train the network. In~\Sref{sec:hyper}, a thorough investigation of the sensitivity to hyperparameters is included.
    
\subsection{Experimental Results}
\label{sec:result}
The results are outlined in~\Tref{tbl:uni_results}, which also compares the proposed method with other state-of-the-art methods. The method of adding the small-loss selection using \Eref{eq:select} to DANCE is called DANCE$_{\text{select}}$. The accuracy under the setting of the pair flipping noise is lower because it is more challenging than the symmetry flipping noise. However, \Tref{tbl:uni_results} demonstrates unequivocally that our method outperformed the existing approaches in all noise levels because they were unable to address all of the challenges of Noisy UniDA. In the Office dataset, the ETN showed acceptable results, and in the VisDA dataset, the TCL achieved excellent performance.
However, our proposed strategy significantly outperformed them. Given the challenging OfficeHome dataset, DANCE performed well when the noise rate was low (e.g., P20 and S20). However, the effectiveness of our approach is comparable to that of DANCE. The proposed method performed better when the noise rate increased (e.g., P45 and S45). Although DANCE$_{\text{select}}$ outperformed the original DANCE in a few settings, our technique still outperformed it in most settings.

The results of each task on each dataset when the noise type was P45 and S45 are shown in~\Tref{tbl:detail_results}. Our approach outperformed in most tasks, proving its ability to identify clean source samples, partially align the distribution, and distinguish the target private classes.

We further demonstrate that the divergence between the two classifiers can separate the common $C$ and target private $\overline{C_t}$ classes in~\Fref{fig:dis}, which shows the probability density function (calculated using kernel density estimation) of the divergence of the common and target private classes in the target domain.
	 
\begin{figure*}[t]
    \begin{minipage}[t]{.32\textwidth}
        \includegraphics[width=\textwidth]{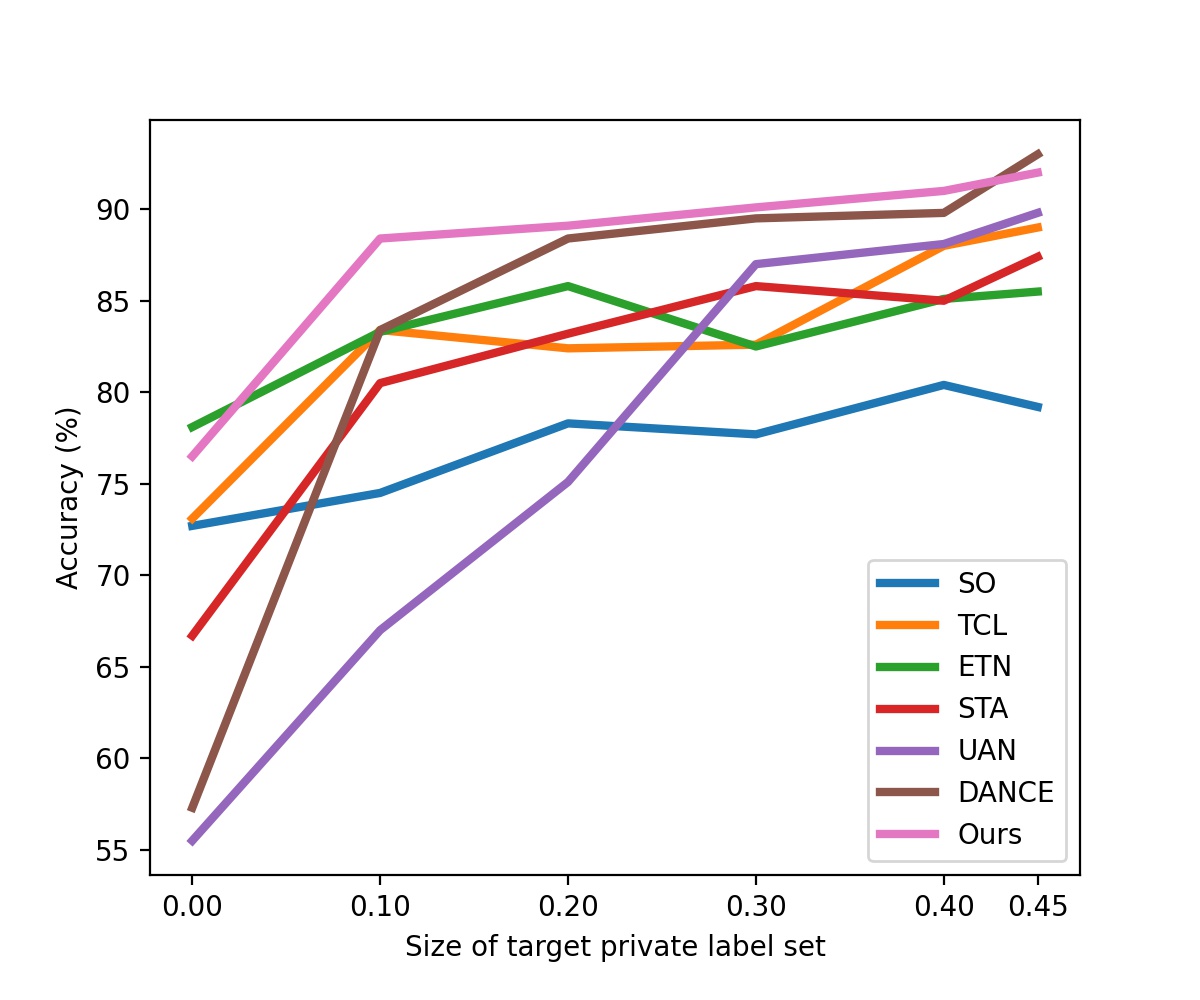}
        \caption{Accuracy w.r.t. $\overline{C_t}$.}
        \label{fig:ab_ct}
    \end{minipage}
    \hfill
    \begin{minipage}[t]{.32\textwidth}
        \includegraphics[width=\textwidth]{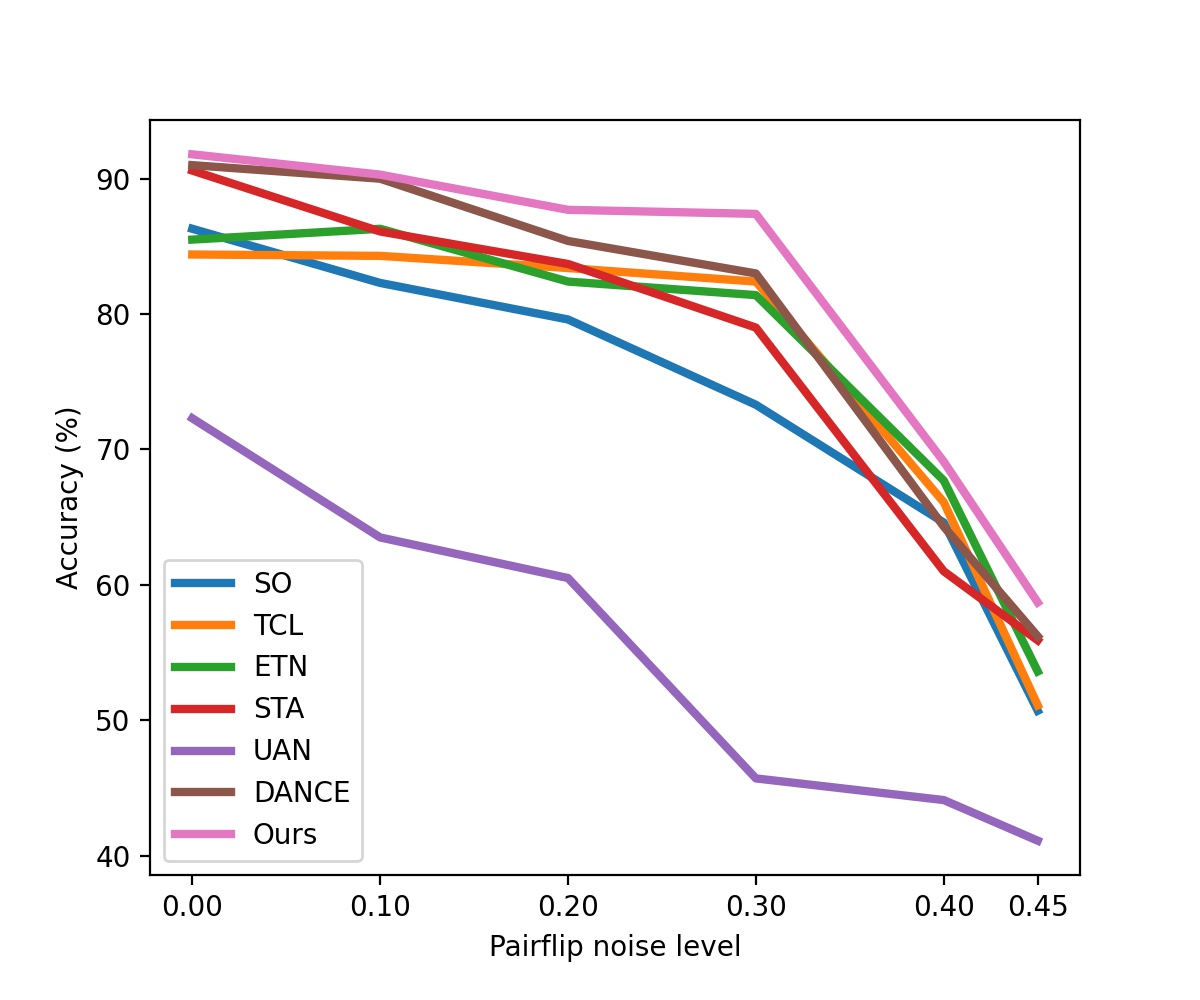}
        \caption{Accuracy w.r.t. pairflip noise level.}
        \label{fig:ab_p}
    \end{minipage}
    \hfill
    \begin{minipage}[t]{.32\textwidth}
        \includegraphics[width=\textwidth]{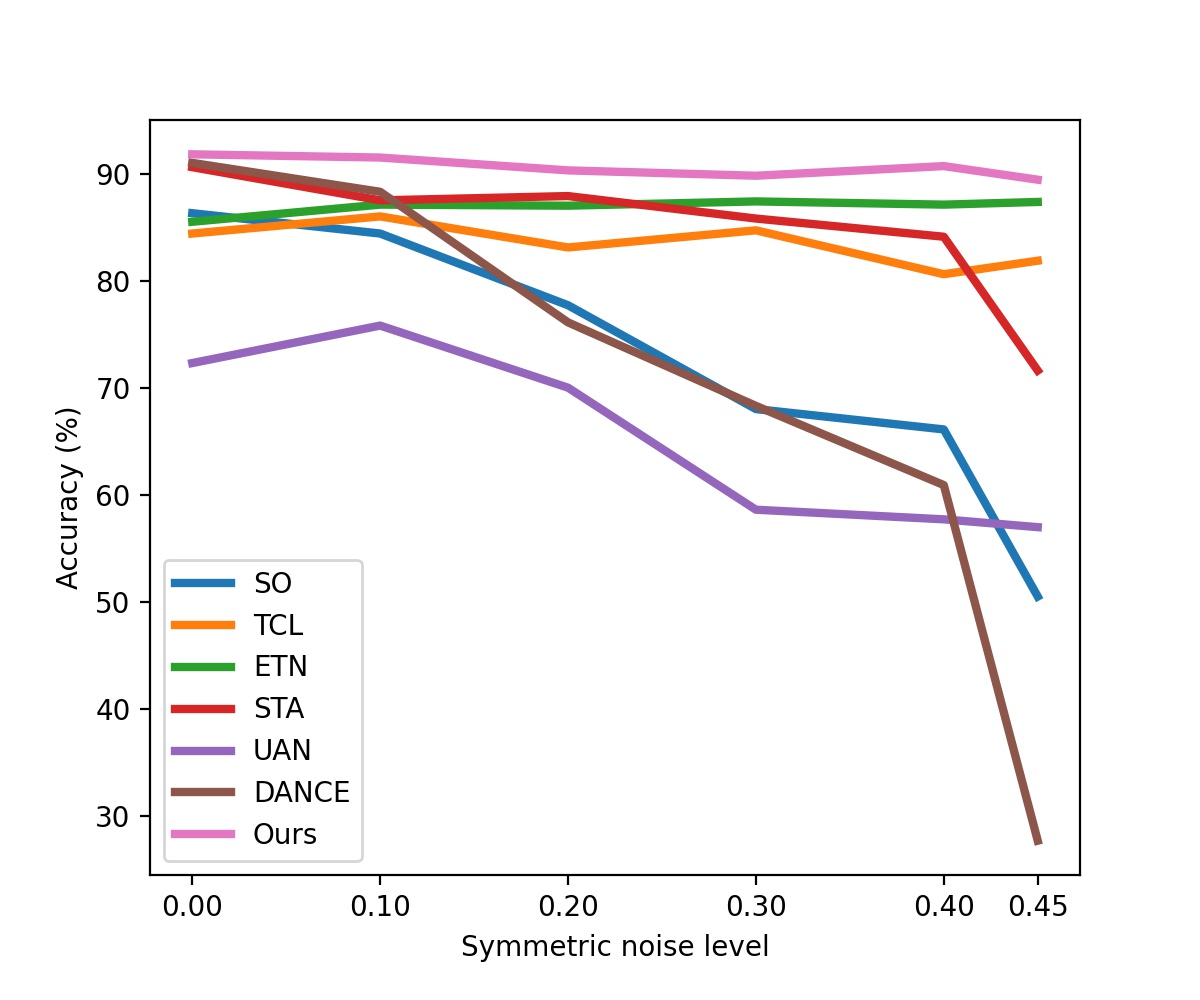}
        \caption{Accuracy w.r.t. symmetric noise level.}
        \label{fig:ab_s}
    \end{minipage}
\end{figure*}

\subsection{Analysis}
\subsubsection{Performance on the Classification of the Source Data}
We divided the source data from the previous domain adaptation experiments into 80\% training data and 20\% test data in order to show how well our strategy handled the noisy label problem. The test data remained clean, whereas the training source data had noisy labels. Even though the evaluation was conducted with source test data, we nonetheless used the ``unknown'' sample rejection technique for all the methods in order to average the accuracy across all classes, including ``unknown'', because we assumed that we do not know the specifics of the category shift. The results are shown in \Tref{tbl:source_results}, and most settings showed that our method performed far better than other methods.

\begin{table*}
\begin{minipage}{0.65\linewidth}
\centering
\caption{Average source-domain accuracy (\%) of each dataset under different noise types. We report the average accuracy of all the tasks for each dataset.}

\tabcolsep = 0.7mm
\begin{tabular}{c|cccc|cccc|cccc}
\toprule
\multirow{2}{*}{Method} & \multicolumn{4}{c|}{Office} & \multicolumn{4}{c|}{OfficeHome} & \multicolumn{4}{c}{VisDA} \\ \cline{2-13}
                        & P20   & P45  & S20  & S45  & P20    & P45   & S20   & S45   & P20  & P45  & S20  & S45  \\ \hline\hline
SO&85.28&57.37&84.19&61.49&78.68&54.49&81.14&59.31&98.69&78.08&77.66&14.03\\
TCL&87.30&55.09&87.05&67.79&77.01&50.36&78.34&61.40&96.89&58.65&97.42&86.18\\
ETN&83.46&57.42&85.09&63.39&80.60&55.45&81.41&61.58&95.53&63.32&97.38&85.76\\
STA&82.26&54.35&82.76&61.17&77.14&50.13&78.57&59.52&94.75&72.58&93.80&85.97\\
UAN&82.77&45.00&84.74&60.75&83.40&51.94&87.92&68.77&96.64&74.42&96.64&95.51\\
DANCE&85.68&58.07&84.20&61.68&80.20&55.53&83.15&62.87&97.62&\textbf{84.55}&61.76&15.15\\
DANCE$_{\text{select}}$&87.35&53.76&88.06&68.10&85.87&54.96&88.97&66.79&97.07&70.75&78.65&85.11\\ \hline
Ours&89.14&\textbf{60.24}&91.07&75.10&83.96&58.30&86.08&76.97&\textbf{98.93}&82.03&\textbf{99.17}&\textbf{98.24}\\
Ours$_{\text{dropout}}$&\textbf{93.91}&59.97&\textbf{92.42}&\textbf{87.54}&\textbf{90.24}&\textbf{62.12}&\textbf{91.12}&\textbf{85.79}&96.79&77.74&95.94&95.15\\\bottomrule
\end{tabular}
\label{tbl:source_results}
\end{minipage}
\begin{minipage}{0.35\linewidth}
\centering
\caption{Ablation study tasks on the Office dataset.}
\tabcolsep = 0.7mm
\begin{tabular}{c|ccc|c}
\toprule
Method & \multicolumn{3}{c|}{Office} & 6 Tasks\\
       & D2W & A2D & W2A & Avg \\ \hline \hline
Ours w/o select     &  94.85  &  88.46 & 85.91  & 86.02 \\
Ours w/o div     &  96.27  &  88.18 & 87.08  & 90.14 \\
Ours w/o crs     &  96.06  & 86.88 & 89.63  & 90.30 \\
Ours w/o ent     &  96.30  &  87.60 & 86.30 & 90.45 \\
Ours w/o sep    &  94.78  & 84.89 &  86.71 &  88.53 \\
Ours w/o mini-max   &  96.05  & 88.71 & 81.03  & 87.31 \\
Ours w/ KL    &  96.19  & 86.49 &  87.77 &  89.75 \\
Ours   &  \textbf{96.65}  & \textbf{89.42} & \textbf{90.90} &  \textbf{91.22} \\ \bottomrule
\end{tabular}
\label{tbl:ab}
\end{minipage}
\end{table*}

\subsubsection{Varying Sizes of $\overline{C_s}$ and $\overline{C_t}$}
We fixed $C_s \cup C_t$ and $C$ in the task A$\rightarrow$D with noise type P20 in the Office dataset and adjusted the sizes of $\overline{C_t}$ to see how well our method performed in various noisy UniDA scenarios. $\overline{C_s}$ was also modified by $\overline{C_s}=C_s \cup C_t \setminus C \setminus \overline{C_t}$.
The comparison of our approach and other methods on various sizes of $\overline{C_t}$ is shown in \Fref{fig:ab_ct}. The performance of our method is comparable to that of ETN, a partial domain adaptation method, for $|\overline{C_t}|=0$, which is the partial domain adaptation setting with $C_t \subset C_s$. The performance of our method is comparable to that of DANCE when $|\overline{C_t}|=21$, which is the open-set domain adaptation setting with $C_s \subset C_t$. Our method performs significantly better than the previous methods in the range of 0 and 21, when $C_s$ and $C_t$ are partially shared. Due to the noisy source samples, the general UniDA methods, UAN and DANCE, performed badly when $|\overline{C_t}|=0$ (partial DA).

\subsubsection{Varying Noise Levels}
We also looked into how label noise affected performance. On task A$\rightarrow$D in the Office dataset, we raised the pairflip and symmetric noise levels from zero to 0.45. \Fref{fig:ab_p} and \Fref{fig:ab_s} show the results, and our method performed well in all settings. When the noise level is zero, it has been found that our method is comparable to the state-of-the-art UniDA method (DANCE). This suggests that all the labels of the source samples are clean. 
Considering the pairflip noise, label noise considerably reduced performance, particularly at high noise levels. Our method nevertheless outperformed other approaches. Considering the symmetric noise, our method was robust to label noise even while the performance of other methods dropped as the noise level rose.

\subsubsection{Ablation Study}
Additionally, we evaluated the variations of our method on the Office dataset with P20 noise to see how effective our method was. (1) Ours w/o select is the variant in which the small-loss selection by \Eref{eq:select} is not deployed. (2) Ours w/o div is the variant in which the divergence component in the classification of the source samples in \Eref{eq:Ls} is not deployed. (3) Ours w/o crs is the variant in which the cross-entropy $\mathcal{\tilde{L}}_{crs}$ between the classifiers of the target samples in \Eref{eq:Lt} is not deployed. (4) Ours w/o ent is the variant in which the entropy $\mathcal{\tilde{L}}_{ent}$ of each classifier of target samples in \Eref{eq:Lt} is not deployed. (5) Ours w/o sep is a variant in which the separation of the divergence between the classifiers as \Eref{eq:Lt} is not deployed. (6) Ours w/o mini-max is the variant in which the mini-max training of the generator and classifiers in \Eref{eq:step_b} and \Eref{eq:step_c} is not deployed to achieve domain alignment. (7) Ours w/ KL is the variant in which the general symmetric KL-divergence in \Eref{eq:Lt} is not deployed. As a result, our approach performed better than other variants in all the settings, as shown in \Tref{tbl:ab}.

\begin{figure*}[t]
    \centering
  \subfloat[Accuracy w.r.t. value of $\alpha$ in \Eref{eq:select}.\label{fig:ab_alpha}]{%
       \includegraphics[width=0.33\linewidth]{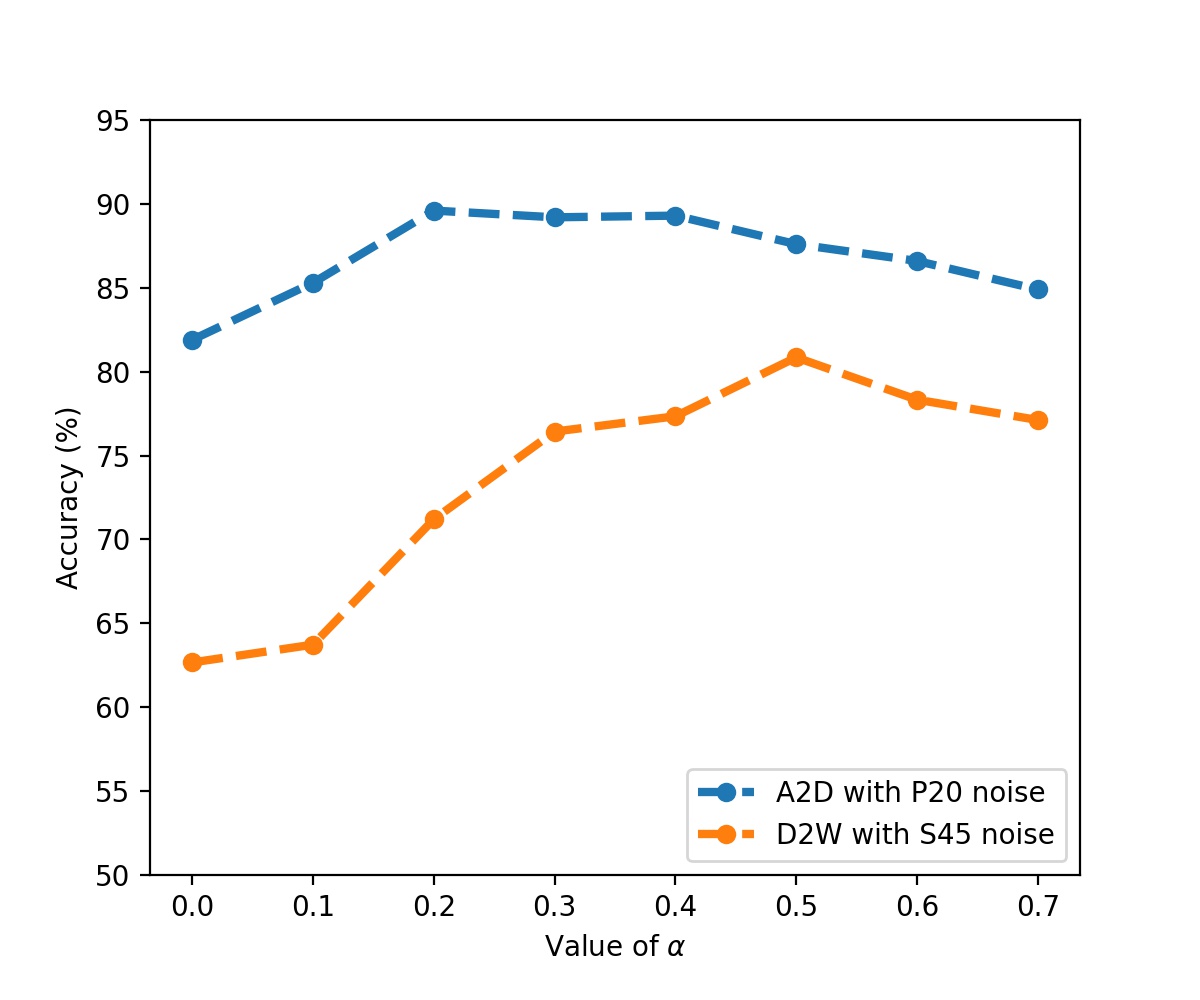}}
    \hfill
  \subfloat[Accuracy w.r.t. value of $\lambda$ in \Eref{eq:Ls}.\label{fig:ab_lambda}]{%
       \includegraphics[width=0.33\linewidth]{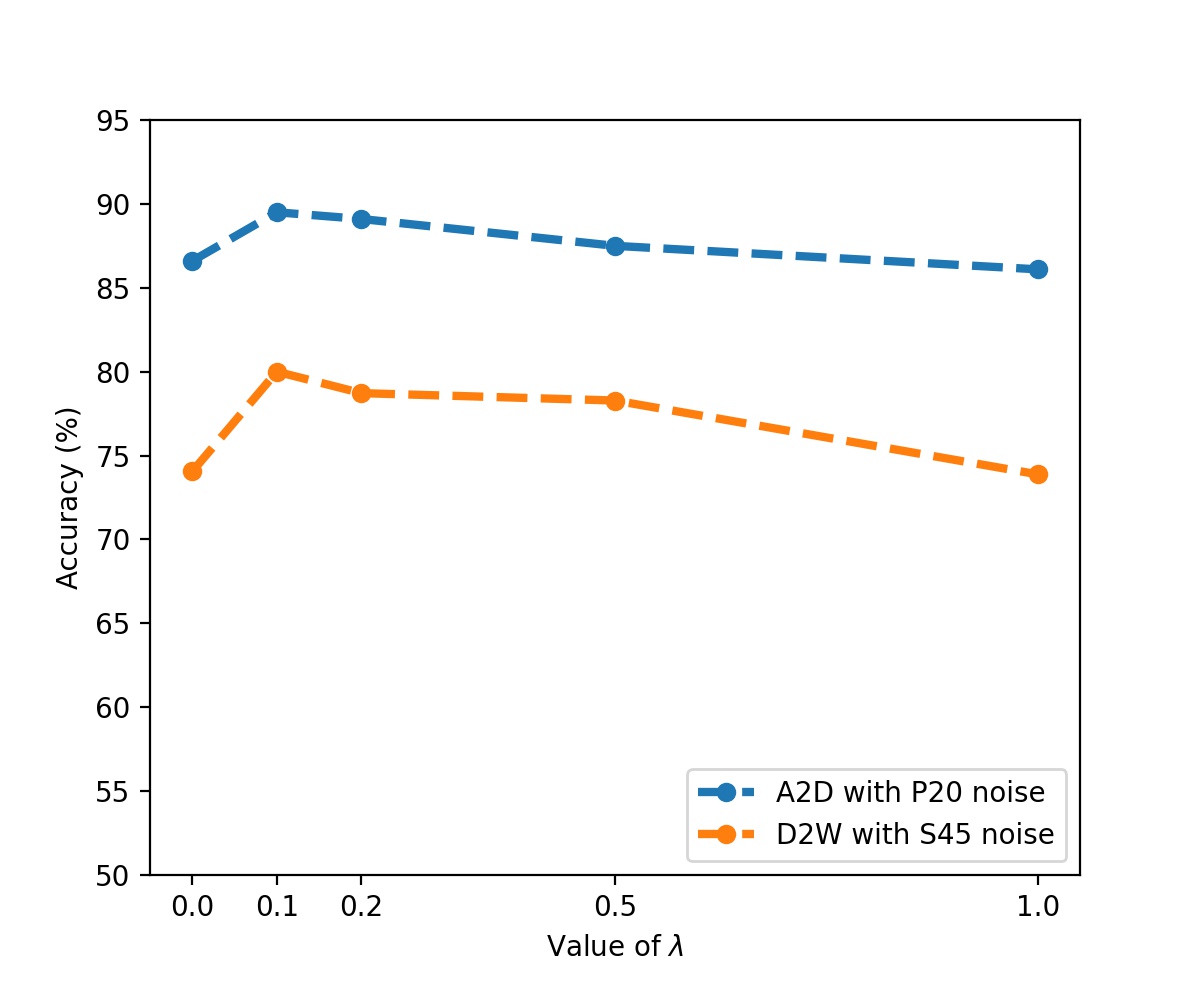}}
    \hfill
  \subfloat[Accuracy w.r.t. value of $\delta$ in \Eref{eq:Lt}.\label{fig:ab_delta}]{%
       \includegraphics[width=0.33\linewidth]{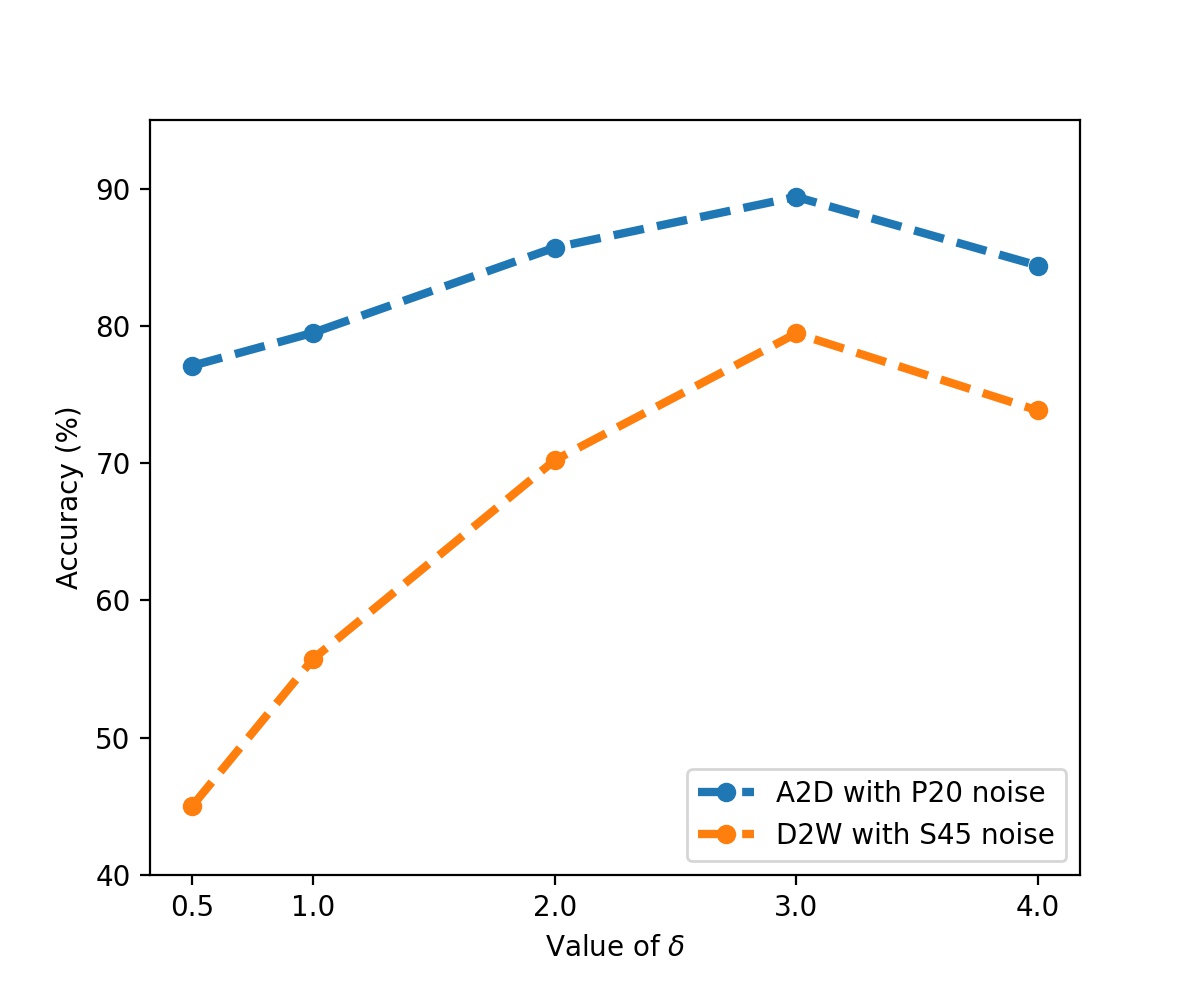}}
    \hfill
  \subfloat[Accuracy w.r.t. value of $m$ in \Eref{eq:Lt}.\label{fig:ab_m}]{%
       \includegraphics[width=0.33\linewidth]{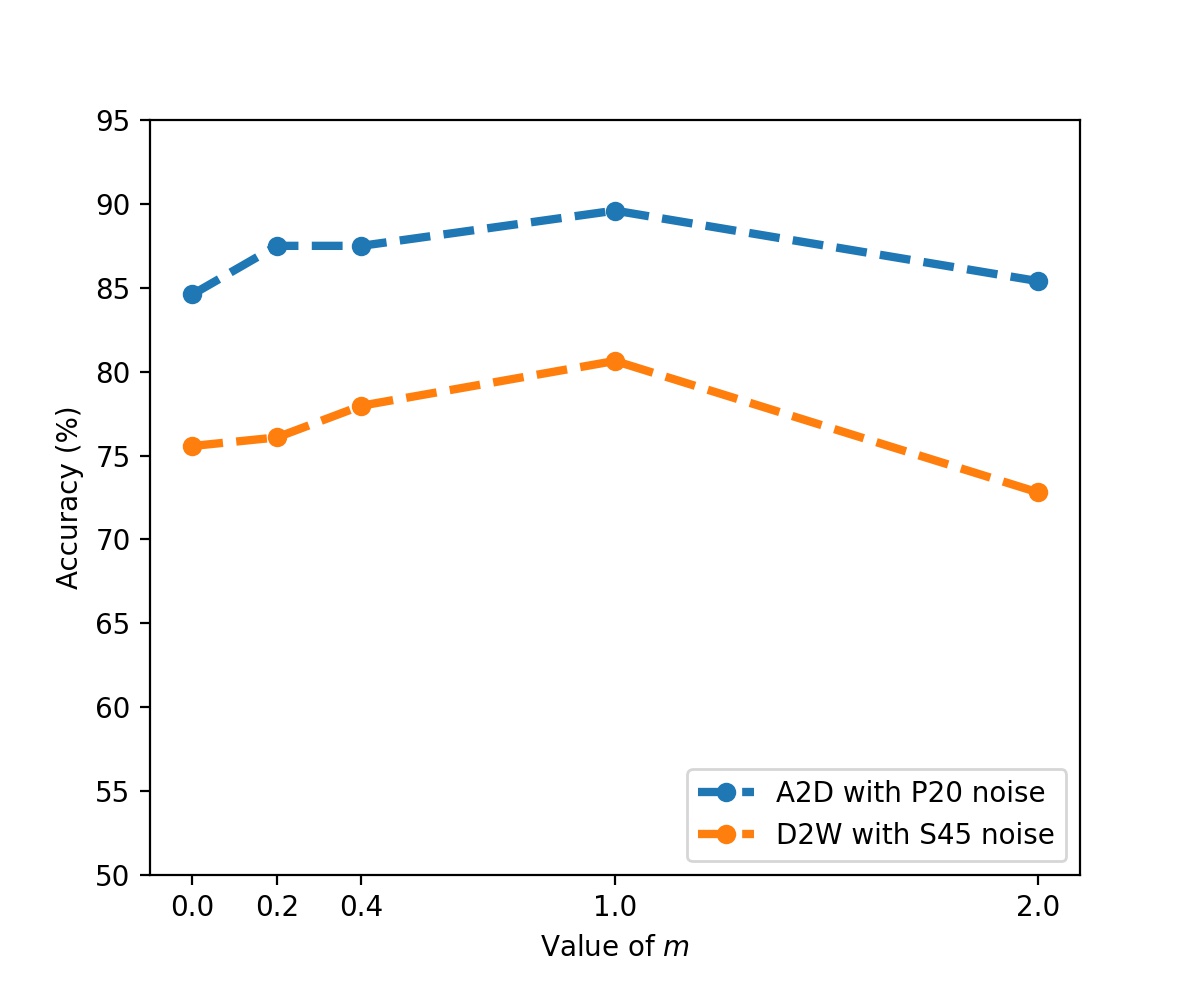}}
  \hspace*{1pt}
  \subfloat[Accuracy w.r.t. value of $n$ in Step C.\label{fig:ab_n}]{%
       \includegraphics[width=0.33\linewidth]{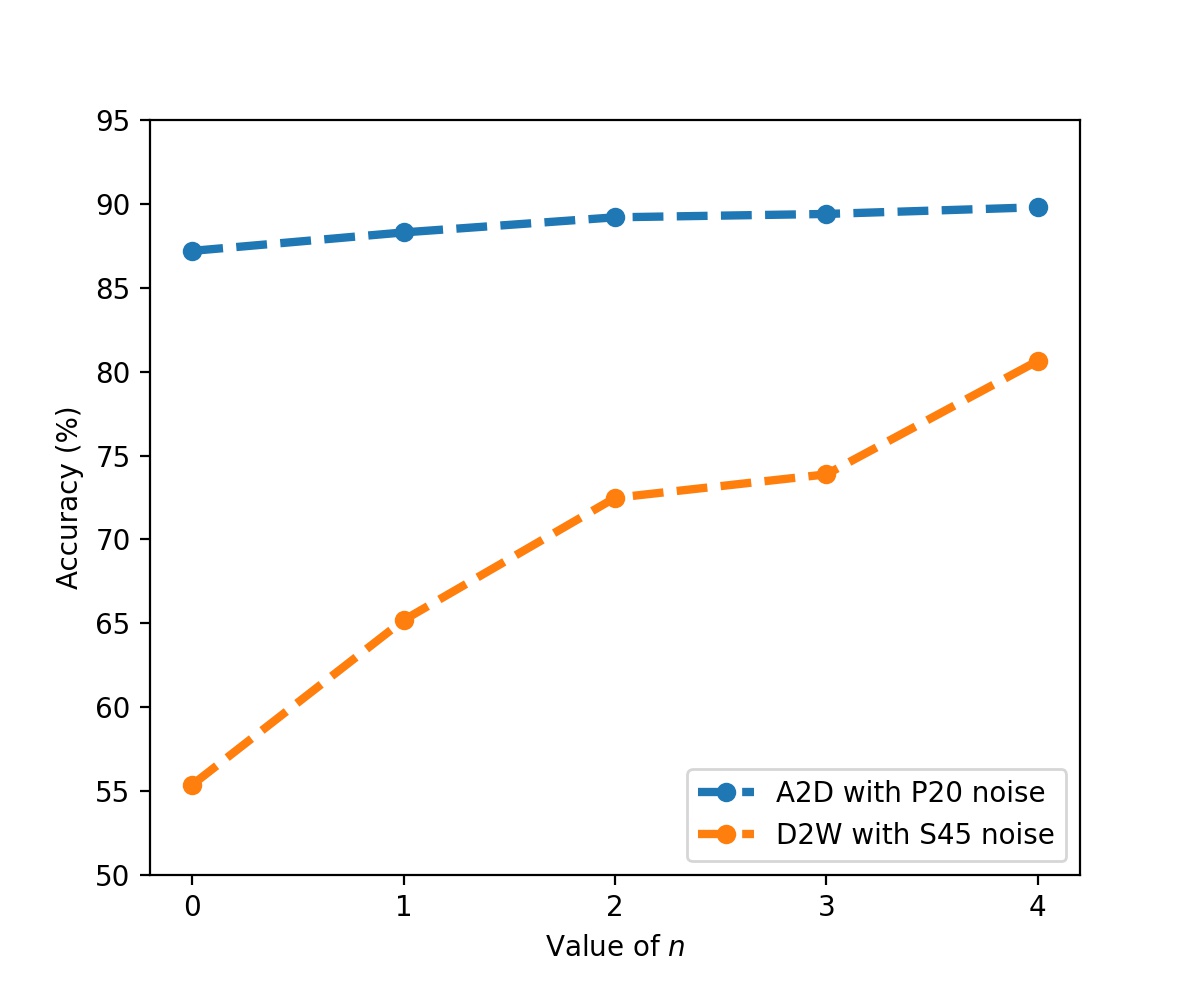}}
  \caption{Analysis of the sensitivity of our method to hyperparameters. }
\label{fig:ab_hp}	
\end{figure*}

\begin{table*}[t]
\centering
\tabcolsep = 1.3mm
\caption{Results on semantic segmentation. The other $\overline{C_t}$ represent the target private classes of ApolloScape (tricycle, traffic cone, road pile, dustbin, and billboard). The bold values represent the highest accuracy for each row.}
\begin{tabular}{c|c|cccccccccccccccc|c}
\toprule
\multirow{2}{*}{Method} & \multicolumn{1}{c|}{\multirow{2}{*}{mIOU}} & \multicolumn{16}{c|}{$C$} & $\overline{C_t}$ \\
 & & Road & sdwk & bldng & Wall & Fence & Pole & Light & Sign & Person & Rider & Car & Truck & Bus & Mcycl & Bcycl & Vgttn & Others \\ \hline\hline
SO & 28.69 &46.61&31.99&25.82&1.56&36.40&19.25&14.51&42.54&17.39&7.48&\textbf{67.23}&31.10&28.33&13.99&32.68&69.53&1.35  \\
DANCE & 27.84 &46.84&36.76&25.82&5.60&38.97&19.45&14.99&42.77&17.21&4.03&65.08&23.64&27.09&5.66&30.38&68.25&0.80  \\
MaxSquare & 29.53&47.23&34.87&24.93&4.06&47.00&17.40&11.23&40.61&12.16&3.52&60.04&\textbf{49.44}&\textbf{41.56}&14.09&27.91&65.92&0.02 \\
Ours & 31.98&47.00&\textbf{37.46}&26.83&7.99&45.19&\textbf{20.30}&\textbf{15.89}&43.33&\textbf{18.70}&9.30&65.84&43.14&36.26&15.26&\textbf{35.38}&70.98&4.86  \\
Ours$_{\text{dropout}}$ & \textbf{32.46}&\textbf{47.49}&35.57&\textbf{27.54}&\textbf{8.05}&\textbf{49.18}&20.29&15.05&\textbf{43.60}&18.41&\textbf{10.97}&63.91&42.80&40.70&\textbf{17.13}&34.14&\textbf{71.86}&\textbf{5.17}\\
\bottomrule
\end{tabular}
\label{tbl:ss_results}
\end{table*}

\subsubsection{Sensitivity to Hyperparameters}
\label{sec:hyper}
We illustrate the sensitivity of our method to hyperparameters on the task A$\rightarrow$D with P20 noise and D$\rightarrow$W with S45 noise in the Office dataset, as shown in \Fref{fig:ab_hp}. $\alpha$ regulates the number of source samples disregarded in each mini-batch. We found that when no sample is dropped ($\alpha=0$), the performance can suffer due to label noise.  Despite the fact that the true noise rate is between 0.2 and 0.45, good performance can still be attained by discarding more samples with greater $\alpha$. The best outcomes are seen in both settings when $\lambda=0.1$, which regulates the weight of the divergence loss on the source samples.
Regarding $\delta$ and $m$, which regulate divergence separation on the target samples, determining $\delta$ as $\log |C_s| \approx 3$ works well and achieving the optimum performance requires setting $m$ around $1$. For $n$ in the training procedure, the number of times Step C is repeated, $n>1$ yields similar results in the P20 A$\rightarrow$D configuration, while $n=4$ yields the best results in the S45 D$\rightarrow$W setting.

\section{Experiments on Semantic Segmentation}
To further demonstrate the effectiveness of our method, we built a new scenario of UniDA for semantic segmentation with real-world label noise and evaluated the proposed method in this setting. Owing to the high annotation cost for semantic segmentation datasets, it is important to achieve adaptation between the different domains in the semantic segmentation.

\subsection{Datasets}
We used the publicly available Cityscapes \cite{cordts2016cityscapes} as the source domain dataset and ApolloScape \cite{huang2019apolloscape} as that of the target domain. To simulate the noisy source situation, we used coarse annotations of the Cityscapes dataset as the training labels, which can represent the label noise in the real world. Because the labels of the Cityscapes and ApolloScape datasets are different, we extracted 16 classes from both datasets as common classes ($C$ in \Tref{tbl:ss_results}), three classes from Cityscapes as source private classes (sky, terrain, and train) and five classes from ApolloScape (tricycle, traffic cone, road pile, dustbin, and billboard) as target private classes.

\subsection{Implementation Detail}
During the training, we randomly sampled two images with their labels of the source dataset and the other two images of the target dataset; however, they did not include labels. We used DRN-D-105 \cite{yu2017dilated} as the segmentation network to evaluate all the methods. Our model was optimized using Momentum SGD with a momentum and learning rate of 0.9 and 0.01 in all the experiments, respectively. The image size was resized to 512 × 512 pixels, and we report the results after 20,000 iterations.

\subsection{Results}
\Tref{tbl:ss_results} shows the comparison of Source Only, DANCE, MaxSquare \cite{Chen_2019_ICCV} (a DA method for semantic segmentation), and our method on the Noisy UniDA of semantic segmentation. These results illustrate that our method is capable of improving the performance. Whereas some methods perform worse than the model trained only on source samples owing to the negative transfer, we can confirm the effectiveness of our method.

\section{Conclusion}
In this study, we propose a framework with divergence optimization for Noisy UniDA. By the use of multiple classifiers, our method can find clean source samples, reject target private classes, and find important target samples that contribute most to the model's adaptation simultaneity. The proposed method greatly outperformed the state-of-the-art methods in multiple benchmarks across numerous source and target domain pairs. The increase in the computational cost is the limitation of the proposed method, because each mini-batch needs to be input into the network several times to update the generator and the classifiers separately. Different from the existing methods for domain adaptation, the proposed method can achieve domain adaptation when some source samples are not correctly labeled. It is especially important for some applications like FoodLog App \cite{aizawa2015foodlog}, which needs to train classification and detection models with low qualities labels collected from users.

\bibliographystyle{IEEEtran}
\bibliography{egbib}

\end{document}